%% file: iclr2027_conference.tex
\documentclass{article} % For LaTeX2e
\usepackage{iclr2027_conference,times}

\input{math_commands.tex}

\usepackage{url}

\usepackage{graphicx}
\usepackage{booktabs}
\usepackage{multirow}
\usepackage{wrapfig}
\usepackage[table]{xcolor}
\usepackage{makecell}     % 关键：之前缺这个导致报错。
\usepackage[most]{tcolorbox}
\usepackage{subcaption}
\usepackage{tabularx}
\usepackage{array}

\definecolor{headblue}{RGB}{178,223,219}
\definecolor{subpurple}{RGB}{225,210,238}
\definecolor{PromptTitleBlue}{RGB}{20, 40, 120}
\definecolor{PromptContentBlue}{RGB}{235, 240, 255}

\newcommand{\bench}{\textsc{EC}\textsuperscript{2}\textsc{-Bench}}
\newcommand{\benchlong}{\textbf{E}dge–\textbf{C}loud \textbf{C}ollaborative Agent Benchmark}

\definecolor{mycitecolor}{RGB}{1,103,204}  % dark navy blue
\definecolor{mylinkcolor}{rgb}{0.55,0.15,0.15}
\usepackage[colorlinks=true,
            citecolor=mycitecolor,   % citations in dark navy blue
            linkcolor=mylinkcolor,         % keep cross-references black
            urlcolor=black]          % keep URLs black
           {hyperref}

\title{Empowering Edge Agents: A Systematic Analysis of Edge-Cloud Collaboration in LLM Agent Execution}

\iclrfinalcopy

\author{Pengcheng Huang$^{1}$, Zhenghao Liu$^{1}$\thanks{ \ \ indicates corresponding author.}, Shixin Ren$^{2}$, Shi Yu$^{2}$, Sijia Yao$^{3}$, \\ \textbf{Yukun Yan$^{2}$\footnotemark[1], Chaojun Xiao$^{2}$\footnotemark[1], Maosong Sun$^{2}$} \\ 
$^1$School of Computer Science and Engineering, Northeastern University, Shenyang, China \\
$^2$Department of Computer Science and Technology, Tsinghua University, Beijing, China\\
$^3$China Mobile Online Services Co., Ltd. Beijing, China\\
}

\begin{document}

\maketitle

\input{sections/0_abs}

\input{sections/1_intro}

\input{sections/2_related_work}

\input{sections/3_bench}

\input{sections/4_eval}

\input{sections/5_ex}
% \vspace{-1.5em}

\input{sections/6_conclusion}
% \input{sections/7_limi}

\bibliographystyle{iclr2027_conference}
\bibliography{iclr2027_conference}

\appendix

\input{sections/n_appendix}

\end{document}

%% file: math_commands.tex
\usepackage{amsmath,amsfonts,bm}

\def\eqref#1{equation~\ref{#1}}
\def\1{\bm{1}}

\DeclareMathAlphabet{\mathsfit}{\encodingdefault}{\sfdefault}{m}{sl}
\SetMathAlphabet{\mathsfit}{bold}{\encodingdefault}{\sfdefault}{bx}{n}

%% file: sections/0_abs.tex
\begin{abstract}

Edge--cloud collaboration offers a promising way to combine the lower inference overhead and greater data locality of edge-side execution with the substantially stronger capabilities of cloud-hosted frontier models. However, as LLM agents increasingly operate in long-horizon scenarios, cloud participation becomes interleaved throughout execution rather than a one-shot decision for each query. Such interleaved participation introduces trajectory-level effects that remain insufficiently understood. To address this gap, we introduce EC\textsuperscript{2}-Bench, a benchmark for evaluating edge--cloud collaboration in long-horizon agent execution. EC\textsuperscript{2}-Bench provides executable digital-workspace agent tasks and a unified evaluation framework covering task performance, resource usage, and privacy exposure. Using this benchmark, we empirically study diverse forms of edge--cloud collaboration. Our study reveals several key findings. First, edge--cloud collaboration provides an effective execution paradigm for long-horizon agents by achieving a favorable balance among task performance, resource usage, and privacy exposure compared with edge-only and cloud-only execution. Second, as edge models become stronger, collaborative execution approaches cloud-only performance, and performance differences across collaboration strategies become smaller, while privacy exposure remains strongly dependent on the chosen strategy. Third, stronger edge capability changes the preferred collaboration mode, shifting from externally controlled cloud participation and direct cloud execution toward more edge-controlled collaboration. Further analyses identify cloud-invocation decision-making and instruction following as important edge-side abilities underlying such effective collaboration. These findings establish a systematic understanding of edge--cloud collaboration in long-horizon agent execution and provide insights for designing future edge models that can effectively collaborate with cloud models. All codes and resources are available at \url{https://github.com/OpenBMB/EC2-Bench}.

\end{abstract}

% agentic help?
% routing,接管类型
% help,帮助类型

%% file: sections/1_intro.tex
% 为什么一定要在agent场景下测端云?
% 首先我们要先说清楚agent场景是重要的, 然后要说清楚agent场景下的端云和之前有什么区别

\section{Introduction}
\label{sec:intro}

% --------------------------------------------- Split ---------------------------------------------
% 按照rq的思路重写-0902 version

Recent advances in large language models (LLMs)~\citep{achiam2023gpt}, together with the emergence of agent harnesses such as Claude Code~\citep{anthropic2025claudecode} and OpenClaw~\citep{openclaw2025}, have shifted LLM applications from isolated prompt--response interactions toward long-horizon, goal-directed agent execution~\citep{wang2024survey}. These agents tackle complex end-to-end workflows, including software development, data analysis, information seeking, and workplace automation, through iterative interaction with external environments and tool use~\citep{yao2022react,yang2024swe,hu2024infiagent,shen2023hugginggpt,zhou2024webarena,wang2024officebench}.

% These agents carry out complex tasks through iterative interaction with external environments, invoking tools and APIs and adapting subsequent decisions based on intermediate outcomes~\citep{yao2022react}. This execution paradigm has enabled agents to tackle increasingly complex end-to-end workflows, including software development, data analysis, information seeking, and workplace automation~\citep{yang2024swe,hu2024infiagent,shen2023hugginggpt,zhou2024webarena,wang2024officebench}.

% v2修改
Bringing such agents to edge devices, however, remains challenging because edge platforms are inherently constrained by limited computational resources and memory~\citep{qu2025mobile}. Consequently, practical on-device deployment often relies on relatively small LLMs (e.g., models with $\leq$10B parameters), which can struggle with demanding reasoning, planning, and tool-use tasks~\citep{choi2026can}, making reliable long-horizon execution challenging~\citep{liu2024mobilellm}. To address this limitation, recent work has explored edge--cloud LLM collaboration, where lightweight edge models are complemented by cloud-hosted frontier LLMs when additional capability is required~\citep{zhang2026hera}. This paradigm combines efficient edge-side execution with stronger cloud capabilities, but requires effective coordination of edge and cloud participation to realize these complementary benefits~\citep{ding2024hybrid,sun2024adaswitch,agrawalllm}.
% This paradigm combines the lower inference overhead and greater data locality of edge-side execution with the substantially stronger capabilities of cloud-hosted frontier models. Realizing these complementary benefits, however, requires effectively coordinating edge and cloud participation during execution~\citep{ding2024hybrid,sun2024adaswitch}.

% As edge models become more capable, a larger share of agent execution can potentially be handled locally, with cloud assistance reserved for operations that require stronger capabilities.
Prior edge--cloud LLM systems largely treated collaboration as a one-shot coordination process for each query, typically through routing or escalation decisions~\citep{pozi2025data,han2026ec}. In long-horizon agent execution, however, collaboration can become more interleaved, with edge and cloud models participating at different points along the trajectory as capability requirements change~\citep{yi2025ecoagent,zhang2026hera}. Such interleaving introduces trajectory-level dependencies: cloud participation decisions can influence subsequent execution steps, accumulate resource costs over the trajectory, and determine how much potentially sensitive information leaves the local environment. Yet existing analysis still largely focuses on isolated routing or escalation decisions at the query level, leaving these trajectory-level effects underexplored~\citep{hu2024routerbench,lu2025routerarena,li2026efficient,li2025collaborative}. To close this gap, in this paper, we conduct a systematic empirical study of edge--cloud collaboration in long-horizon agentic execution. Specifically, we organize our study around three research questions. We first ask: \emph{does edge--cloud collaboration remain effective for realistic long-horizon agent execution?} (\textbf{RQ1}) We evaluate this question jointly in terms of task performance, resource usage, and privacy exposure. As edge models become more capable, more execution can be handled locally while cloud assistance is reserved for capability-demanding operations. We therefore ask: \emph{how does increasing edge capability impact edge--cloud collaborative execution?} (\textbf{RQ2}); As increasing edge capability changes how computation is distributed between edge and cloud, collaboration strategies that are effective under one capability regime may not remain equally suitable under another. This motivates our final question: \emph{how should edge--cloud collaboration evolve as edge capability advances?} (\textbf{RQ3}).

% todo: 这里加上我们bench独特的端云设计, 更好的证明,就得用我们bench测端云.=》必要性
To provide a common empirical basis for this study, we introduce \bench{} (\benchlong{}), an evaluation framework that combines a diverse suite of real-world digital-workspace agent tasks with a unified execution and evaluation environment for edge--cloud collaboration. \bench{} characterizes each configuration along three dimensions: task performance~\citep{multi2026device}, resource usage~\citep{fang2025bridging}, and privacy exposure. We explicitly consider privacy because cloud participation may reveal sensitive workspace information encountered during agent execution, making privacy exposure an important dimension of evaluating such collaboration~\citep{fu2026ci}. Using \bench{}, we systematically investigate edge--cloud collaboration across diverse agent tasks, collaboration strategies, and edge-capability regimes.

Our study yields three main findings. First, edge--cloud collaboration remains effective for long-horizon agents: partial cloud participation improves task performance over edge-only execution while requiring substantially less cloud resource usage and privacy exposure than cloud-only execution. Second, as edge models become stronger, collaborative execution approaches cloud-only performance, and performance differences across collaboration strategies become much smaller; however, their differences in privacy exposure remain substantial. Thus, privacy remains an important differentiator even when task performance converges. Third, the collaboration strategy that works best changes with edge capability. Weaker edge models benefit more from external routing and direct cloud execution, whereas stronger edge models benefit more from edge-controlled invocation and cloud advisory. Further analyses suggest that effective cloud-invocation decisions and instruction following are key edge-side abilities for stronger-edge collaboration modes. These findings highlight the value of optimizing future edge models for both standalone capability and effective cloud collaboration, enabling more capable edge agents. 

% Further analyses suggest that effective cloud-invocation decisions and instruction following are two important edge-side abilities for these stronger-edge collaboration modes. These findings highlight a promising direction for future edge models: improving not only their standalone capabilities, but also their ability to effectively collaborate with cloud models to extend the capabilities of edge agents.

%% file: sections/2_related_work.tex
\section{Related Work}

% ------------------------------------------ split 0901version ------------------------------------------
% 这个 section 也要从“谁和我们的 benchmark 最像”转成“已有工作分别告诉了我们什么、还缺什么 understanding”。
A growing body of work has investigated how edge and cloud LLMs can be coordinated to leverage their complementary capabilities and deployment characteristics, balancing overall task performance with local efficiency~\citep{zhu2025lslm,dong2025hybridflow}. A substantial line of work focuses on one-shot coordination for individual queries, either by routing each query to an edge or cloud model or by cascading from the local model to the cloud when additional capability is needed~\citep{chen2023frugalgpt,ong2024routellm,miaoselective}. These approaches therefore center on improving the accuracy of routing and escalation decisions~\citep{ding2024hybrid}. More recent work has extended edge--cloud collaboration to agent execution, where edge and cloud participation can be coordinated at different steps throughout a multi-step trajectory rather than determined once for the entire query, enabling finer-grained use of their complementary capabilities. Within this setting, recent approaches have explored different designs for coordinating edge--cloud collaboration, mainly along two dimensions. Some dynamically determine when the cloud model should participate, either through dedicated rule-based controllers or routers, or by allowing the edge agent itself to decide when to invoke the cloud~\citep{shao2025division,zhang2026hera,zhang2026router,zeng2026learning}. Beyond deciding when the cloud should participate, other approaches have explored complementary functional roles for edge and cloud models during execution, such as execution, planning, or providing guidance~\citep{yi2025ecoagent,yuan2026paac}.
% Motivated by this diversity, our focus is not on the effectiveness of any single coordination mechanism, but on the broader behavior of edge--cloud collaboration as an execution paradigm for agents.

Existing work has begun to develop dedicated evaluation resources for edge--cloud LLM systems. LLMSwitchBench~\citep{liu2025llmswitchbench} evaluates edge--cloud routing for smart-home queries, while M4A1~\citep{multi2026device} extends device--cloud evaluation to multimodal, multitask, and multi-turn conversations, focusing on where individual inference requests should be executed. More general LLM routing benchmarks instead abstract away from the edge--cloud setting and evaluate routing systems over heterogeneous model pools, primarily focusing on routing quality and overall cost--performance trade-offs~\citep{huang2025routereval,lu2025routerarena,li2026llmrouterbench}. Despite this progress, existing evaluations pay more attention on routing decisions for individual queries, tasks, or turns, typically over standalone inference workloads such as knowledge, reasoning, coding, and instruction-following tasks. Our study instead treats edge--cloud collaboration in long-horizon agent execution as the primary object of empirical analysis, aiming to systematically characterize its effectiveness and system-level behavior across different edge capability levels and collaboration designs.

%% file: sections/3_bench.tex
\section{Preliminaries: Edge--Cloud Collaborative Execution}
\label{subsec:collaborative_execution}
\begin{figure}[t]
    \centering
    \includegraphics[width=\linewidth]{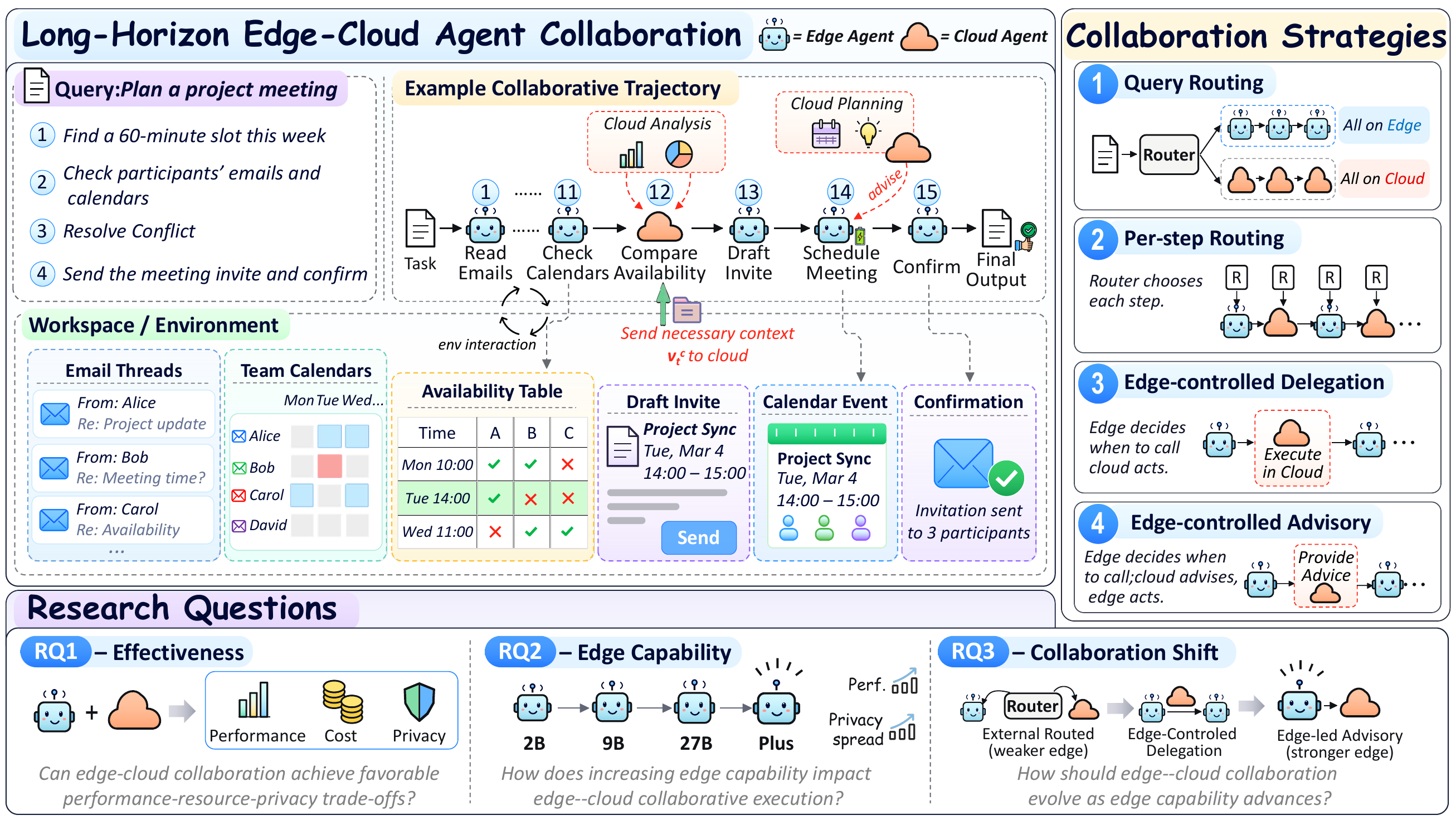}
    % \vspace{-1em}
    \caption{
Overview of edge--cloud collaboration for long-horizon agent execution in \bench{}. The figure illustrates a representative agent workflow, four collaboration modes with different forms of cloud participation, and three research questions on collaboration effectiveness, the impact of edge capability, and how collaboration shifts as edge models become stronger.
}
    \vspace{-1.2em}
    \label{fig:overview}
\end{figure}

This section formalizes edge--cloud collaboration in agent execution, which serves as the object of study in this work. We first present a general formulation of how a collaboration strategy coordinates the participation of edge and cloud models during execution (\S\ref{subsec:collaboration_formulation}), and then instantiate this formulation with four representative collaboration strategies examined in our experiments (\S\ref{subsec:collaboration_strategies}). Figure~\ref{fig:overview} provides an intuitive overview of edge--cloud collaborative agent execution.
% Figure~\ref{fig:overview} provides an intuitive overview of this setting.

\subsection{Edge--Cloud Collaboration Formulation}
\label{subsec:collaboration_formulation}

Edge--cloud collaboration considers a system comprising an edge model $M_e$ and a remotely hosted cloud model $M_c$, together with a collaboration strategy $\mathcal{C}$ that determines how computation is distributed between them during agent execution. The edge model operates within the local execution boundary using a lightweight LLM (typically $\leq$10B parameters), whereas the cloud model offers substantially stronger capabilities at the cost of greater inference overhead, enabling collaboration to balance task capability with the efficiency and data locality of edge-side execution.

Given a task instruction $I_{\mathrm{Task}}$, execution under $\mathcal C$ proceeds through iterative interactions with the environment. At interaction step $t$, let
\begin{equation}
\small
\tau_{<t}
=
(a_1,o_1,\ldots,a_{t-1},o_{t-1}),
\end{equation}
denote the partial execution trajectory, where $a_i$ is the agent action and $o_i$ is the resulting environment observation. Based on $I_{\mathrm{Task}}$ and the partial execution trajectory $\tau_{<t}$, $\mathcal C$ determines the participating model(s) at the current step through a selection rule $\pi_{\mathcal C}$:
\begin{equation}
\small
\mathcal M_t
\leftarrow
\pi_{\mathcal C}
\left(
I_{\mathrm{Task}},\tau_{<t}
\right),
\qquad
\mathcal M_t
\in
\big\{
\{M_e\},
\{M_c\},
\{M_e,M_c\}
\big\}.
\end{equation}
Depending on $\mathcal C$, this selection may be fixed before execution or determined dynamically as execution proceeds. The next agent action $a_t$ is then generated by the participating model(s) according to $\mathcal C$. Once $a_t$ is generated, it is executed in the environment to produce the observation:
\begin{equation}
\small
o_t=\operatorname{Env}(a_t).
\end{equation}
The resulting pair $(a_t,o_t)$ is then appended to the execution trajectory $\tau_{<t}$. Repeating this process until termination yields the complete execution trajectory $\tau_T=(a_1,o_1,\ldots,a_T,o_T)$. To characterize how collaboration unfolds beyond the agent--environment trajectory, we additionally record the participating models and the cloud-visible context $v_t^c$ at each step. Specifically, if $M_c\in\mathcal M_t$, $v_t^c$ denotes the context provided to the cloud model at step $t$ for generating its output, including task information, execution history, and other context specified by $\mathcal C$; otherwise, $v_t^c=\emptyset$. This yields the collaborative execution trace:
\begin{equation}
\small
\gamma_{\mathcal C}
=
\left(
\tau_T,
(\mathcal M_1,v_1^c),
\ldots,
(\mathcal M_T,v_T^c)
\right).
\end{equation}
% To characterize how collaboration unfolds beyond the agent--environment trajectory, we additionally record the participating models and the cloud-visible context at each step. Specifically, if $M_c\in\mathcal M_t$, the cloud-visible context $v_t^c$ denotes the execution information exposed to the cloud model at step $t$; otherwise, $v_t^c=\emptyset$. This yields the collaborative execution trace:

% Depending on $\mathcal C$, this selection may be fixed before execution or determined dynamically as execution proceeds. If $M_c\in\mathcal M_t$, the cloud model receives a cloud-visible context $v_t^c$ containing the execution information exposed to it at the current step; otherwise, no cloud interaction occurs and $v_t^c=\emptyset$. The next agent action $a_t$ is then generated through the participating model(s) according to $\mathcal C$. Once $a_t$ is generated, it is executed in the environment to produce the observation:
% \begin{equation}
% \small
% o_t=\operatorname{Env}(a_t).
% \end{equation} The resulting pair $(a_t,o_t)$ is then appended to the execution trajectory $\tau_{<t}$. Repeating this process until termination yields the complete execution trajectory $\tau_T=(a_1,o_1,\ldots,a_T,o_T)$. To characterize collaborative execution beyond the agent--environment trajectory, we additionally record model participation and cloud-visible context at each step, yielding the collaborative execution trace:
% \begin{equation}
% \small
% \gamma_{\mathcal C}
% =
% \left(
% \tau_T,
% (\mathcal M_1,v_1^c),
% \ldots,
% (\mathcal M_T,v_T^c)
% \right).
% \end{equation}

\subsection{Collaboration Strategies}
\label{subsec:collaboration_strategies}

Building on the formulation above, we consider four representative collaboration strategies that instantiate $\mathcal C$ and represent common forms of edge--cloud collaboration explored in prior work~\citep{ong2024routellm,zeng2026glimprouter,farzaneh2026think,anthropic2026advisor}.

% query routing
\textit{Query Routing.}
Query Routing employs an external router to select the model for a task before agent execution begins. Routers typically use a lightweight scoring or classification model to estimate, from the task instruction $I_{\mathrm{Task}}$, whether the task should be handled by the edge model $M_e$ or the stronger cloud model $M_c$~\citep{ong2024routellm}. The routing decision is made once and remains fixed throughout execution, such that $\mathcal M_t$ remains either $\{M_e\}$ or $\{M_c\}$ at every execution step, and the entire trajectory is generated by the selected model.

% Per-step routing
\textit{Per-step Routing.} 
Per-step Routing revisits model selection at each interaction step. Instead of making a single task-level decision, an external router estimates, based on the task instruction $I_{\mathrm{Task}}$ and the partial execution trajectory $\tau_{<t}$, whether the current step can be handled by the edge model $M_e$ or should be routed to the stronger cloud model $M_c$~\citep{zeng2026glimprouter}. The routing decision is updated throughout execution, with $\mathcal M_t$ switching between $\{M_e\}$ and $\{M_c\}$ across steps, thereby enabling finer-grained and dynamic use of cloud capabilities.

% adaptive edge delegation
\textit{Edge-Controlled Cloud Delegation.} 
Edge-Controlled Cloud Delegation shifts the participation decision from an external router to the edge agent itself, motivated by the intuition that an executing agent, with direct access to its evolving execution state, may better identify when cloud assistance is needed~\citep{farzaneh2026think}. Specifically, execution remains on the edge by default, i.e., $\mathcal M_t=\{M_e\}$, while a dedicated invocation tool allows the agent to delegate the current step to the cloud when stronger model capabilities are required. Upon invocation, $\mathcal M_t$ switches to $\{M_c\}$, and the cloud model generates the next action before control returns to the edge agent.

% On-Demand Cloud advisory
\textit{Edge-Controlled Cloud Advisory.}
Edge-Controlled Cloud Advisory follows the same agent-controlled invocation paradigm but uses the cloud model in an advisory role rather than delegating action generation~\citep{anthropic2026advisor}. Execution remains edge-led by default, with $\mathcal M_t=\{M_e\}$. When the edge agent requests cloud assistance, the cloud model participates in the current step, yielding $\mathcal M_t=\{M_e,M_c\}$, and provides auxiliary information such as plans, suggestions, or reasoning. The edge model then incorporates this guidance to make a better-informed decision while retaining responsibility for generating the next action.

% \section{\bench{}: Benchmarking Edge--Cloud Collaborative Execution}
\section{\bench{}: Benchmarking Edge--Cloud Collaboration for Long-Horizon Agent Execution}
\label{sec:benchmark}

\input{tables/bench_diff}

To systematically study edge--cloud collaboration in agent execution, we introduce \bench{}, comprising 128 executable agent tasks and a unified procedure for evaluating different collaboration strategies. For each task, we execute the agent under a collaboration strategy $\mathcal{C}$ and record the resulting collaborative execution trace $\gamma_{\mathcal C}$, which captures both the agent trajectory and cloud participation throughout execution. \bench{} then evaluates the resulting execution along three dimensions: task performance, resource usage, and privacy exposure, providing a comprehensive view of the effectiveness and costs of edge--cloud collaboration. These design choices distinguish \bench{} from existing benchmarks for LLM routing and edge--cloud collaboration. Table~\ref{tab:benchmark_comparison} summarizes the key differences among these benchmarks in evaluation scenarios, collaboration modes, and evaluation dimensions, with a more detailed comparison provided in Appendix~\ref{app:benchmark_comparison}.

\subsection{Task Suite Construction}
\label{subsec:benchmark_tasks}
We construct \bench{} by collecting executable agent tasks from existing benchmarks, curating them for the study of edge--cloud collaboration, and augmenting the resulting task suite with sensitivity annotations for privacy-exposure evaluation.

\textbf{Data Collection.}
We construct the task suite of \bench{} based on six established executable agent benchmarks: Claw-Eval~\citep{ye2026claw}, WildClawBench~\citep{ding2026wildclawbench}, QwenClawBench~\citep{qwenclawbench1.1}, LiveClawBench~\citep{long2026liveclawbench}, PinchBench~\citep{pinchbench2026}, and ClawBench~\citep{zhang2026clawbench}. We select these benchmarks because their tasks evaluate end-to-end agent execution through stateful, multi-step interactions with tools, applications, and workspace artifacts, with explicit task objectives and verifiable execution outcomes. Collectively, they cover a broad range of applications, interaction patterns, execution environments, and task difficulties, providing a diverse candidate pool for constructing \bench{}.

\textbf{Task Curation.}
Starting from the collected candidate pool, we further adapt and curate the tasks to construct the final task suite. Specifically, we first adapt tasks from different source benchmarks to a common execution environment and evaluation interface, enabling all tasks to be executed under the same agent runtime and evaluated under consistent conditions while preserving their original objectives, workspace states, required interactions, and success criteria. We then apply a collaboration-oriented difficulty filtering procedure to retain tasks that are informative for studying edge--cloud collaboration. We evaluate each adapted task under edge-only and cloud-only execution to characterize task difficulty at the two execution endpoints. Tasks that are consistently solved by both endpoints are removed because they provide limited opportunity to differentiate agent execution paradigms, while tasks where neither endpoint achieves meaningful completion are excluded because they provide limited basis for evaluating collaborative execution. Following these steps, we further supplement the task suite with tasks from underrepresented scenarios to maintain broad workload coverage. The resulting task suite consists of 128 executable tasks.
% Tasks that are consistently solved by both endpoints are removed because they leave little room for cloud assistance, while tasks that neither endpoint can solve reliably are excluded because they provide little basis for evaluating the benefit of cloud assistance. Following these steps, we supplement a small number of tasks from underrepresented scenarios to maintain broad workload coverage. This process yields 128 executable tasks.

To measure the potential exposure of sensitive information from local workspaces to the cloud during collaboration, we further incorporate sensitivity annotations into \bench{}. Specifically, we first annotate sensitive information already contained in the original workspace artifacts. We then use an LLM-based approach to identify appropriate locations, generate realistic synthetic sensitive information, and insert it into relevant workspace artifacts, while preserving the original task objective, required interactions, and evaluation criteria. This process results in fine-grained sensitivity annotations for 100 tasks, where each sensitivity unit corresponds to an atomic piece of sensitive information grounded in a workspace artifact.

% To measure the potential exposure of sensitive information from local workspaces to the cloud during collaboration, we further incorporate sensitivity annotations into \bench{}. Specifically, we construct fine-grained sensitivity annotations for 100 tasks, where each sensitivity unit corresponds to an atomic piece of sensitive information grounded in a workspace artifact. We annotate sensitive information already present in the original tasks and, where appropriate, introduce realistic synthetic sensitive content while preserving the task objective, required interactions, and evaluation criteria.

Detailed procedures for task adaptation, collaboration-oriented filtering, coverage supplementation, and sensitivity annotation are provided in Appendix~\ref{app:task_statistics}, Appendix~\ref{app:task_adaptation} and Appendix~\ref{app:privacy_annotation_construction}.

\subsection{Evaluating Collaborative Execution}
\label{subsec:benchmark_evaluation}

Given the collaborative execution trace $\gamma_{\mathcal C}$ produced by a collaboration strategy $\mathcal C$, \bench{} evaluates the execution along three dimensions: task performance, resource usage, and privacy exposure. Task performance measures whether the agent accomplishes the task, resource usage quantifies the computation consumed by the edge and cloud models during execution, and privacy exposure measures the fraction of annotated sensitive information units that become visible to the cloud during collaborative execution.
% privacy exposure captures the extent to which sensitive local information is revealed to the cloud. 
% Together, these dimensions provide a comprehensive view of the task outcome and the resource and privacy implications of a collaboration strategy.

\textbf{Task Performance.}
We assess task performance using task-specific evaluation checks that cover key requirements for successful completion, including whether the intended outcome is achieved, task constraints are satisfied, and required tool or application interactions are performed correctly. Checks that can be determined directly from task outputs or workspace states are evaluated through rule-based validation or environment-state auditing, while checks requiring semantic judgment are evaluated using rubric-based LLM-as-a-judge~\citep{ding2026wildclawbench}. Given an execution trajectory $\tau$ contained in $\gamma_{\mathcal C}$, each check produces a score in $[0,1]$, and the task completion score is computed as the weighted average of these check scores according to predefined task-specific weights.
% We assess task performance using task-specific evaluation checks defined for each task. Each task specifies a set of checks that assess different aspects of successful completion, including whether the intended task outcome is achieved, task-specific constraints are satisfied, and required tool or application interactions are performed correctly. Checks that can be determined directly from task outputs or workspace states are evaluated through rule-based validation or environment-state auditing, while checks requiring semantic or qualitative judgment are evaluated using rubric-based LLM-as-a-judge. Given an execution trajectory $\tau$ contained in $\gamma_{\mathcal C}$, each check produces a score in $[0,1]$, and the task completion score is computed as the weighted average of these check scores according to their predefined task-specific weights.

\textbf{Resource Usage.}
We quantify the resource usage of collaborative execution from both the cloud and edge sides. For the cloud, we record the input, output, and cache-read tokens consumed by all cloud invocations in $\gamma_{\mathcal C}$, and compute the corresponding monetary cost according to the pricing of the cloud model. For the edge, we estimate the total floating-point operations (FLOPs) incurred by the edge model throughout the execution trajectory.

\textbf{Privacy Exposure.}
We measure privacy exposure by quantifying how much sensitive information becomes visible to the cloud during collaborative execution. For each task with sensitivity annotations, let $\mathcal{S}$ denote its set of sensitivity units. Given the cloud-visible contexts $\{v_t^c\}_{t=1}^{T}$ recorded in $\gamma_{\mathcal C}$, we measure privacy exposure as the fraction of sensitivity units disclosed to the cloud:
\begin{equation}
E_{\mathrm{priv}}
=
\frac{1}{|\mathcal{S}|}
\sum_{s\in\mathcal{S}}
\mathbb{I}_{\mathrm{exp}}
\left(
s,\{v_t^c\}_{t=1}^{T}
\right),
\label{eq:privacy_exposure}
\end{equation}
where $\mathbb{I}_{\mathrm{exp}}$ indicates whether the information represented by sensitivity unit $s$ is disclosed to the cloud during execution. Since sensitive information may be paraphrased or reformulated rather than reproduced verbatim, we use an LLM-based judge for exposure detection~\citep{zharmagambetov2026agentdam}. To validate the automated judge, we compare its predictions against adjudicated human labels on 200 sensitivity unit--context pairs. The automated judge achieves an F1 score of 96.9\% against human annotations, demonstrating high agreement with human judgments and the reliability of our privacy annotations. Full validation results are reported in Appendix~\ref{app:privacy_leakage_validation}.
% This metric captures the coverage of sensitive information exposed during execution, providing a trajectory-level measure of how much annotated sensitive content becomes accessible to the cloud.

%% file: tables/bench_diff.tex
% \begin{table*}[t]
%     \centering
%     \small
%     \caption{Comparison of \bench{} with prior benchmarks for edge--cloud collaboration and LLM routing. Task Setting denotes the interaction form for task execution. For collaboration modes, $R_q$ and $R_s$ denote query- and step-level routing, respectively, while $D$ and $C$ denote on-demand cloud delegation and consultation. Evaluation summarizes the dimensions considered by each benchmark.}
%     \label{tab:benchmark_comparison}
%     \vspace{-0.5em}
%     \resizebox{\textwidth}{!}{
%     \begin{tabular}{lccc}
%         \toprule
%         \textbf{Benchmark} & \textbf{Task Setting} & \textbf{Collab. Modes} & \textbf{Evaluation Dim.} \\
%         \midrule
%         RouterBench~\citep{hu2024routerbench} & General LLM & $R_q$ & Task Perf. / Cost \\
% LLMRouterBench~\citep{li2026llmrouterbench} & General LLM and Tool-Use & $R_q$ & Task Perf. / Cost \\
% LLMSwitchBench~\citep{liu2025llmswitchbench} & Single-turn Query & $R_q$ & Task Perf. / Cost \\
% M4A1~\citep{multi2026device} & Multi-turn Query & $R_s$ & Task Perf. / Cost / Latency \\
% \midrule
% \bench{} & Long-horizon Agent & $R_q$, $R_s$, $D$, $C$ & Task Perf. / Cost / Privacy \\
%         \bottomrule
%     \end{tabular}}
%     \vspace{-0.8em}
% \end{table*}

\begin{table*}[t]
    \centering
    \small
    \caption{
    Comparison of \bench{} with prior benchmarks for edge--cloud collaboration and LLM routing. \textbf{Evaluation Scenario} specifies the task and interaction setting considered by each benchmark. For \textbf{Collab. Modes}, $R_q$ and $R_s$ denote query-level and per-step routing, respectively, while $D$ and $A$ denote edge-controlled cloud delegation and cloud advisory, respectively. \textbf{Evaluation Dim.} summarizes the primary evaluation metrics considered by each benchmark.
    }
    \label{tab:benchmark_comparison}
    \vspace{-0.5em}
    \resizebox{\textwidth}{!}{
    \begin{tabular}{lccc}
        \toprule
        \textbf{Benchmark} & \textbf{Evaluation Scenario} & \textbf{Collab. Modes} & \textbf{Evaluation Dim.} \\
        \midrule
        RouterBench~\citep{hu2024routerbench} & General LLM Tasks & $R_q$ & Task Perf. / Cost \\
        LLMRouterBench~\citep{li2026llmrouterbench} & General LLM and Tool-Use Tasks & $R_q$ & Task Perf. / Cost \\
        LLMSwitchBench~\citep{liu2025llmswitchbench} & Single-turn Query & $R_q$ & Task Perf. / Cost \\
        M4A1~\citep{multi2026device} & Multi-turn Dialogue & $R_s$ & Task Perf. / Cost / Latency \\
        \midrule
        \bench{} & Long-horizon Agent Execution & $R_q$, $R_s$, $D$, $A$ & Task Perf. / Resource / Privacy \\
        \bottomrule
    \end{tabular}}
    \vspace{-1em}
\end{table*}

%% file: sections/4_eval.tex
% 

\section{Experimental Methodology}
\label{sec:experimental_setup}

This section describes the experimental configurations and evaluation protocol used to study edge--cloud collaboration on \bench{}.

% Qwen3.5
\textbf{Execution Configurations.}
We instantiate the edge side with models of progressively increasing capability from the Qwen series~\citep{qwen35}. Qwen3.5-2B and Qwen3.5-9B serve as representative edge models under current deployment constraints, while Qwen3.5-27B and Qwen3.7-Plus are included as higher-capability proxies for future edge models. We use GLM-5.3~\citep{glm5team2026glm5vibecodingagentic} as the primary cloud model across all main configurations, given its strong performance on agentic tasks. We further evaluate our findings with DeepSeek-V4.1-Flash as an alternative cloud model to examine their robustness beyond a single cloud model (Appendix~\ref{app:deepseek_v41}). For comparison, we include edge-only and cloud-only execution, where action generation is handled exclusively by $M_e$ and $M_c$, respectively, throughout the task. All configurations are implemented within OpenClaw\footnote{\url{https://openclaw.ai/}}, providing a shared agent runtime and tool interface across execution settings.

\textbf{Evaluation Protocol.}
Each task--configuration pair is executed independently three times to account for randomness in agent execution~\citep{chan2025mle}, with results averaged across runs. In addition to the task completion score defined in Section~\ref{subsec:benchmark_evaluation}, we report $\mathrm{Pass}^{3}$ to characterize execution robustness, defined as the fraction of tasks successfully completed in all three runs, where a run is considered successful if its completion score exceeds 0.7. Privacy exposure is evaluated on the 100 tasks equipped with sensitivity annotations. For task checks requiring rubric-based semantic judgment and for privacy-exposure detection, we use GPT-5.4-mini~\citep{openai2026gpt54mini} with reasoning enabled, following prior practice in LLM-based agent evaluation~\citep{zharmagambetov2026agentdam}. Cloud-side monetary cost is computed using OpenRouter\footnote{\url{https://openrouter.ai/}} pricing based on raw input, cache-read input, and output tokens, while edge-side computation is estimated in FLOPs following~\citet{kaplan2020scaling}. Detailed pricing assumptions and FLOPs calculations are provided in Appendix~\ref{app:pricing}.

%% file: sections/5_ex.tex
\input{tables/main_table}
\section{Empirical Study}
We now use \bench{} to systematically study edge--cloud collaboration in long-horizon agent execution. Our analysis follows the three research questions introduced in Section~\ref{sec:intro}, examining whether edge--cloud collaboration is effective, how improving edge capability affects collaborative execution, and how collaboration mechanisms evolve as edge models become stronger.

% \subsection{Does Edge--Cloud Collaboration Work for Agentic Execution? (RQ1)}
\subsection{Effectiveness of Edge--Cloud Collaboration in Agent Execution (RQ1)}

\label{sec:rq1}

\input{figs/rq1}

We first examine whether edge--cloud collaboration provides a useful execution regime between edge-only and cloud-only agents when task performance, resource usage, and privacy exposure are considered jointly. As shown in Table~\ref{tab:main_tab}, the two execution extremes exhibit clear trade-offs. Cloud-only execution achieves the strongest task performance but incurs substantial cloud resource usage and privacy exposure, whereas edge-only execution avoids cloud usage and exposure at the cost of lower task performance. Collaborative execution occupies the space between these extremes, improving task performance over edge-only execution while substantially reducing cloud dependence relative to cloud-only execution. For example, with Qwen3.5-2B as the edge model, step-level routing improves task completion from 12.75 to 39.97 while using only 22.7\% of the cloud-only cost. These results indicate that partial cloud participation can recover substantial capability without incurring the full cost of cloud-only execution.

\input{figs/rq2}

We further examine whether these trade-offs reflect benefits beyond a proportional mixture of the two execution extremes. For each edge model, we use the linear interpolation between its edge-only and cloud-only endpoints as a reference, representing the performance expected from proportionally trading local and cloud execution~\citep{zhengclear}. As shown in Figures~\ref{fig:rq1_a} and~\ref{fig:rq1_b}, multiple collaborative configurations lie above the corresponding interpolation baselines in the cost--performance and privacy--performance spaces, achieving higher task performance at comparable cloud cost or privacy exposure. These results show that edge--cloud collaboration can provide gains beyond an interpolation between edge-only and cloud-only execution.

\subsection{Impact of Improving Edge Capability on Collaborative Execution (RQ2)}
\label{subsec:rq2}

As edge models continue to advance rapidly~\citep{xiao2025densing}, we study how improving edge capability changes edge--cloud collaborative execution. To probe how collaboration may evolve beyond current lightweight edge models, we additionally include Qwen3.5-27B and Qwen3.7-Plus as prospective proxies for stronger future edge models. As shown in Table~\ref{tab:main_tab}, stronger edge models consistently improve collaborative execution and bring its performance closer to cloud-only execution. The best available collaborative configuration increases from 39.97\% completion with Qwen3.5-2B to 76.84\% with Qwen3.7-Plus, approaching the 82.70\% completion of cloud-only execution.

We further examine how increasing edge capability changes the variation among collaboration strategies. Specifically, for each edge model, we measure the gap between the highest and lowest completion scores and privacy exposure values. As shown in Figures~\ref{fig:rq2_a} and~\ref{fig:rq2_b}, task performance becomes substantially less sensitive to the collaboration strategy as edge capability increases. The completion gap among the evaluated strategies narrows from 25.22\% with Qwen3.5-2B to 3.67\% with Qwen3.7-Plus. In contrast, privacy exposure does not exhibit the same convergence, with substantial differences persisting across collaboration strategies. Notably, with Qwen3.7-Plus, the strategies differ by only 3.67\% in completion but by 66.23\% in privacy exposure. These results show that stronger edge capability substantially compresses performance differences across collaboration strategies, while privacy exposure remains highly differentiated.

\subsection{How Does Edge--Cloud Collaboration Evolve with Edge Capability? (RQ3)}
\label{subsec:rq3}

Beyond improving collaborative performance, increasing edge capability may also change effective collaboration mechanisms. We examine this shift along two dimensions: who determines cloud participation and what role the cloud plays once invoked. As shown in Table~\ref{tab:main_tab}, both dimensions exhibit clear shifts with improving edge capability. In terms of participation control, externally routed strategies generally outperform edge-controlled strategies with weaker edge models. For example, with Qwen3.5-2B, Per-step Routing achieves 39.97\% completion, compared with 15.86\% for Edge-Ctrl. Delegation. As edge capability improves, however, this pattern reverses. A similar reversal appears in the role of the cloud: weaker edge models benefit more from direct cloud execution, whereas stronger edge models favor advisory support.

\input{figs/figs/rq3}

% To translate this scaling trend into actionable design guidance
% These shifts suggest that Edge-Controlled Cloud Advisory is particularly effective for stronger edge models among the evaluated strategies. To understand the factors associated with this scaling trend, we further examine two edge-side abilities that may contribute to this behavior: cloud-assistance decision making and cloud-guidance utilization. Detailed experimental settings and additional analyses are provided in Appendix~\ref{app:rq3_analysis}.
These shifts motivate further analysis of two edge-side abilities associated with this trend: cloud-assistance decision making and cloud-guidance utilization. Detailed experimental settings and additional analyses are provided in Appendix~\ref{app:rq3_analysis}.

\textit{Cloud-Assistance Decision Making.}
The shift toward agent-controlled invocation motivates us to examine whether stronger edge models can make more effective cloud-assistance decisions. Since there is no explicit annotation indicating when cloud assistance is necessary, we compare Edge-Controlled Cloud Delegation against a matched random-invocation baseline. Specifically, we preserve the same cloud-call rate as the agent-controlled strategy while randomly selecting invocation steps. As shown in Figure~\ref{fig:rq3_a}, edge-controlled delegation consistently outperforms random invocation. The gain is only 1.2\% for Qwen3.5-2B, but reaches 3.0--4.4\% for the more capable edge models. This pattern suggests that stronger edge models may make more effective cloud-invocation decisions.

\textit{Cloud Guidance Utilization.}
We next examine whether instruction-following ability is associated with improved advisory collaboration. Starting from Qwen3.5-2B, we apply targeted instruction-following training and evaluate intermediate checkpoints under a fixed cloud-advisory setting. As shown in Figure~\ref{fig:rq3_b} and Appendix~\ref{app:rq3_analysis}, task completion under cloud advisory consistently improves as instruction-following ability develops during training, increasing from 24.8\% before training to 28.8\% after 600 steps. These results suggest that instruction-following ability is associated with improved utilization of cloud-provided guidance.

% We next examine whether improving the edge model's instruction-following ability is associated with better advisory collaboration. We apply targeted instruction-following training to the edge model and evaluate intermediate checkpoints. As shown in Figure~\ref{fig:rq3_b}, advisory task completion increases consistently with training, from 24.8\% before training to 28.8\% after 600 steps. This result suggests that improving instruction-following ability can enhance the effectiveness of advisory collaboration. 

% These observations suggest that, in addition to general agent capability, cloud-assisted decision-making and guidance following are collaboration-relevant abilities worth considering when developing edge models for edge--cloud systems. 

%% file: tables/main_table.tex
\definecolor{headblue}{HTML}{B7DDD9}
\definecolor{preexec}{HTML}{F4F0FF}
\definecolor{runtime}{HTML}{FFF3E6}

\newcommand{\preexecbox}{\protect\raisebox{0.15ex}{\protect\colorbox{preexec}{\protect\rule{0pt}{0.5em}\protect\rule{0.2em}{0pt}}}}
\newcommand{\runtimebox}{\protect\raisebox{0.15ex}{\protect\colorbox{runtime}{\protect\rule{0pt}{0.5em}\protect\rule{0.2em}{0pt}}}}

\begin{table*}[!t]
\centering
\small
\caption{
Performance of edge--cloud collaboration strategies on \bench{}.
\textbf{Performance}: task completion and Pass$^3$.
\textbf{Resource Usage}: Cloud Tok. (R/C/O) reports raw input, cache-read input, and output tokens in millions; Cloud Cost is reported in USD; Edge FLOPs are reported in PetaFLOPs.
\textbf{Privacy Exposure}: percentage of sensitive information exposed to the cloud (lower is better).
\runtimebox{} and \preexecbox{} denote current-generation and higher-capability proxy edge models, respectively. For Qwen3.7-Plus, Edge FLOPs are not reported because its parameter count is unavailable.
}
\label{tab:main_tab}
% \vspace{-0.5em}
\resizebox{\textwidth}{!}{
\begin{tabular}{lcccccc}
\toprule
\multirow{2}{*}{\textbf{Method}} & \multicolumn{2}{c}{\textbf{Performance$\uparrow$}} & \multicolumn{3}{c}{\textbf{Resource Usage$\downarrow$}} & \multirow{2}{*}{\textbf{Privacy Exp.$\downarrow$}} \\
\cmidrule(lr){2-3} \cmidrule(lr){4-6}
& Completion & Pass$^3$ & Cloud Tok. (R/C/O) & Cloud Cost & Edge FLOPs & \\
\midrule

\rowcolor{headblue}\multicolumn{7}{c}{\textbf{\textit{Edge-only}}} \\
Qwen3.5-2B~(\citeyear{qwen35}) & 12.75 & 2.34 & -- & -- & 142.98 & 0.00 \\
Qwen3.5-9B~(\citeyear{qwen35}) & 39.12 & 9.38 & -- & -- & 121.64 & 0.00 \\
Qwen3.5-27B~(\citeyear{qwen35}) & 61.79 & 30.47 & -- & -- & 564.49 & 0.00 \\
Qwen3.7-Plus~(\citeyear{qwen35}) & 71.34 & 53.12 & -- & -- & -- & 0.00 \\
% Qwen3.8-Max~(\citeyear{qwen35}) & 70.76 & 45.31 & -- & -- & 1542.25 & 0.00 \\
\midrule

\rowcolor{headblue}\multicolumn{7}{c}{\textbf{\textit{Cloud-only}}} \\
DeepSeek-v4-Pro~(\citeyear{deepseekai2026deepseekv4}) & 72.90 & 53.12 & 5.81 / 50.20 / 1.60 & 16.41 & -- & 86.66 \\
GLM-5.3~(\citeyear{openai2026gpt54}) & 82.70 & 71.88 & 13.52 / 65.76 / 4.64 & 56.42 & -- & 87.01 \\
\midrule

\rowcolor{headblue}\multicolumn{7}{c}{\textbf{\textit{Edge-Cloud Collaboration}}} \\

\textit{\textbf{Qwen3.5-2B}} \\
\rowcolor{runtime}\quad Query Routing~(\citeyear{ong2024routellm}) & 23.98 & 12.50 & 1.81 / 10.49 / 0.79 & 8.76 & 156.30 & 12.59 \\
\rowcolor{runtime}\quad Per-step Routing~(\citeyear{zeng2026glimprouter}) & 39.97 & 7.81 & 3.84 / 17.29 / 0.67 & 12.80 & 57.05 & 41.29 \\
\rowcolor{runtime}\quad Edge-Ctrl. Delegation~(\citeyear{farzaneh2026think}) & 15.86 & 2.34 & 0.41 / 0.49 / 0.11 & 1.19 & 172.74 & 2.71 \\
\rowcolor{runtime}\quad Edge-Ctrl. Advisory~(\citeyear{anthropic2026advisor}) & 14.75 & 1.56 & 0.67 / 0.51 / 0.29 & 2.36 & 240.28 & 6.07 \\
\midrule

\textit{\textbf{Qwen3.5-9B}} \\
\rowcolor{runtime}\quad Query Routing~(\citeyear{ong2024routellm}) & 43.60 & 17.97 & 2.13 / 11.28 / 0.78 & 9.33 & 162.88 & 12.59 \\
\rowcolor{runtime}\quad Per-step Routing~(\citeyear{zeng2026glimprouter}) & 52.71 & 21.88 & 1.98 / 7.46 / 0.31 & 6.07 & 29.96 & 27.52 \\
\rowcolor{runtime}\quad Edge-Ctrl. Delegation~(\citeyear{farzaneh2026think}) & 52.88 & 17.19 & 0.83 / 1.03 / 0.36 & 3.01 & 184.98 & 16.45 \\
\rowcolor{runtime}\quad Edge-Ctrl. Advisory~(\citeyear{anthropic2026advisor}) & 60.46 & 26.56 & 2.02 / 2.61 / 1.19 & 8.73 & 332.43 & 45.46 \\
\midrule

\textit{\textbf{Qwen3.5-27B}} \\
\rowcolor{preexec}\quad Query Routing~(\citeyear{ong2024routellm}) & 65.53 & 38.28 & 2.69 / 14.29 / 0.75 & 10.76 & 649.92 & 12.53 \\
\rowcolor{preexec}\quad Per-step Routing~(\citeyear{zeng2026glimprouter}) & 68.42 & 42.19 & 1.76 / 6.51 / 0.26 & 5.29 & 88.58 & 35.81 \\
\rowcolor{preexec}\quad Edge-Ctrl. Delegation~(\citeyear{farzaneh2026think}) & 69.06 & 44.53 & 1.00 / 1.50 / 0.71 & 4.93 & 455.76 & 42.42 \\
\rowcolor{preexec}\quad Edge-Ctrl. Advisory~(\citeyear{anthropic2026advisor}) & 70.13 & 47.66 & 1.32 / 0.75 / 0.83 & 5.71 & 822.23 & 48.46 \\
\midrule

\textit{\textbf{Qwen3.7-Plus}} \\
\rowcolor{preexec}\quad Query Routing~(\citeyear{ong2024routellm}) & 73.17 & 60.16 & 4.39 / 15.60 / 0.88 & 14.03 & -- & 12.51 \\
\rowcolor{preexec}\quad Per-step Routing~(\citeyear{zeng2026glimprouter}) & 73.77 & 56.25 & 2.17 / 4.95 / 0.22 & 5.28 & -- & 53.17 \\
\rowcolor{preexec}\quad Edge-Ctrl. Delegation~(\citeyear{farzaneh2026think}) & 74.24 & 60.94 & 1.65 / 1.08 / 0.33 & 4.03 & -- & 43.36 \\
\rowcolor{preexec}\quad Edge-Ctrl. Advisory~(\citeyear{anthropic2026advisor}) & 76.84 & 67.19 & 1.97 / 0.53 / 0.88 & 6.80 & -- & 78.74 \\

% \textit{\textbf{Qwen3.8-Max}} \\
% \rowcolor{preexec}\quad Query Routing~(\citeyear{ong2024routellm}) & 80.03 & 70.31 & 2.31 / 11.23 / 0.75 & 9.43 & 2180.89 & 12.46 \\
% \rowcolor{preexec}\quad Per-step Routing~(\citeyear{zeng2026glimprouter}) & -- & -- & -- & -- & -- & -- \\
% \rowcolor{preexec}\quad Edge-Ctrl. Delegation~(\citeyear{farzaneh2026think}) & 80.63 & 71.88 & 2.74 / 3.52 / 1.02 & 9.22 & 2171.17 & 80.89 \\
% \rowcolor{preexec}\quad Edge-Ctrl. Advisory~(\citeyear{anthropic2026advisor}) & 82.64 & 72.66 & 2.73 / 1.00 / 0.80 & 7.61 & 2315.85 & 77.54 \\

\bottomrule
\end{tabular}
}

% \vspace{-1.3em}
\end{table*}

%% file: figs/rq1.tex
% Model colors
% \definecolor{qwenblue}{HTML}{56B4E9}
% \definecolor{maxmagenta}{HTML}{CC79A7}
% \definecolor{glmorange}{HTML}{E69F00}

\definecolor{qwenblue}{HTML}{009E73}
\definecolor{maxmagenta}{HTML}{0072B2}
\definecolor{glmorange}{HTML}{E69F00} % 如果 GLM 还是这个颜色的话

% Colored boxes used in captions
\newcommand{\qwenbox}{%
    \raisebox{0.15ex}{\textcolor{qwenblue}{\rule{0.9ex}{0.9ex}}}%
}
\newcommand{\maxbox}{%
    \raisebox{0.15ex}{\textcolor{maxmagenta}{\rule{0.9ex}{0.9ex}}}%
}
\newcommand{\glmbox}{%
    \raisebox{0.15ex}{\textcolor{glmorange}{\rule{0.9ex}{0.9ex}}}%
}

% \begin{figure}[t]
\begin{wrapfigure}{r}{0.47\columnwidth}
    \centering
    \vspace{-1em}
    \begin{subfigure}[t]{0.485\linewidth}
        \centering
        \includegraphics[width=\linewidth]{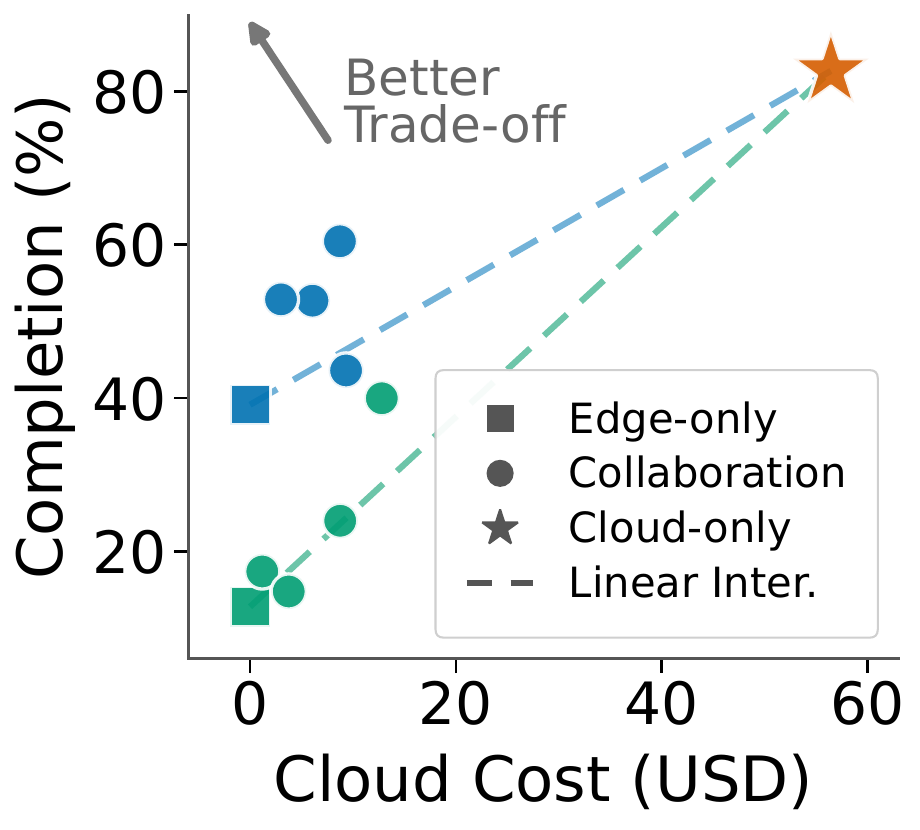}
        \caption{Cost-performance.}
        \label{fig:rq1_a}
    \end{subfigure}
    \hfill
    \begin{subfigure}[t]{0.485\linewidth}
        \centering
        \includegraphics[width=\linewidth]{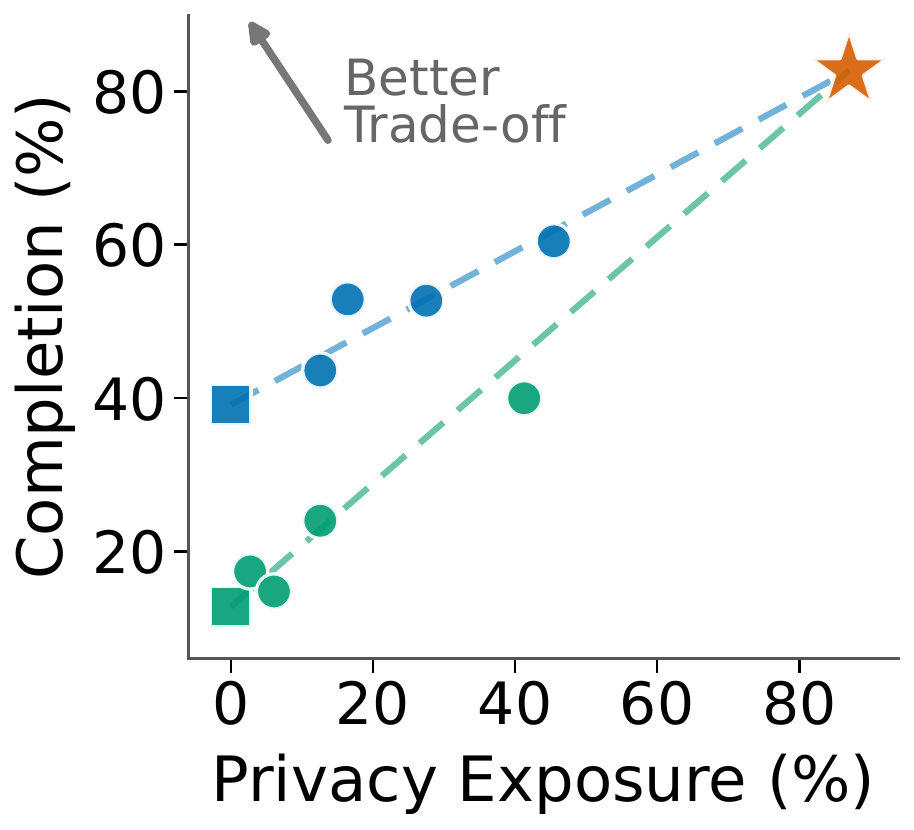}
        \caption{Privacy-performance.}
        \label{fig:rq1_b}
    \end{subfigure}

    % \caption{
    %     Trade-offs of different execution modes.
    %     (a) Cost--performance trade-off.
    %     (b) Privacy--performance trade-off.
    %     \qwenbox{} and \maxbox{} denote Qwen3.5-2B and Qwen3.5-9B, respectively.
    %     Each circle represents a different edge--cloud collaboration mode.
    %     In both plots, points closer to the upper-left indicate more favorable
    %     trade-offs, achieving higher task completion with lower cloud cost or
    %     privacy exposure. Dashed lines denote the linear interpolation between
    %     edge-only and cloud-only execution. Points above the dashed lines achieve
    %     higher task utility than the interpolation baseline at the same cloud cost
    %     or privacy exposure.
    % }
    
\caption{ Trade-offs of edge--cloud execution modes. \qwenbox{} and \maxbox{} denote Qwen3.5-2B and Qwen3.5-9B. Each circle represents a collaboration mode; upper-left is better. Dashed lines interpolate between edge-only and cloud-only execution, with points above the lines indicating gains beyond the interpolation baseline. }
    \label{fig:rq1}
    \vspace{-1em}
\end{wrapfigure}

%% file: figs/rq2.tex
% Model colors
% \definecolor{qwenblue}{HTML}{56B4E9}
% \definecolor{maxmagenta}{HTML}{CC79A7}
% \definecolor{glmorange}{HTML}{E69F00}

\definecolor{qwenblue}{HTML}{009E73}
\definecolor{maxmagenta}{HTML}{0072B2}
\definecolor{glmorange}{HTML}{E69F00} % 如果 GLM 还是这个颜色的话

\begin{wrapfigure}{r}{0.48\columnwidth}
    \centering
    \vspace{-1.3em}
    \begin{subfigure}[t]{0.48\linewidth}
        \centering
        \includegraphics[width=\linewidth]{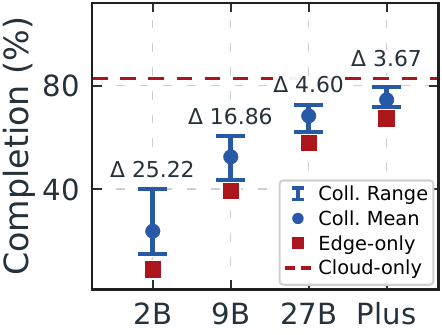}
        \caption{Task completion.}
        \label{fig:rq2_a}
    \end{subfigure}
    \hfill
    \begin{subfigure}[t]{0.48\linewidth}
        \centering
        \includegraphics[width=\linewidth]{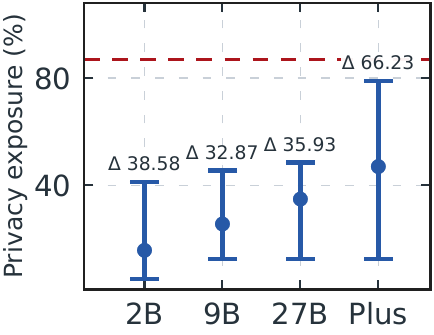}
        \caption{Privacy exposure.}
        \label{fig:rq2_b}
    \end{subfigure}
    
    \caption{
    Task completion and privacy exposure across edge-capability regimes. Blue points denote the mean values across edge--cloud collaboration configurations, and vertical bars indicate the range across configurations. The x-axis labels abbreviate the corresponding edge models (Qwen3.5-2B/9B/27B and Qwen3.7-Plus).
    }
    \label{fig:rq2}
    \vspace{-1.4em}
\end{wrapfigure}

%% file: figs/figs/rq3.tex
\begin{wrapfigure}{r}{0.42\columnwidth}
    \centering
    \vspace{-1em}
    \begin{subfigure}[t]{0.5\linewidth}
        \centering
        \includegraphics[width=\linewidth]{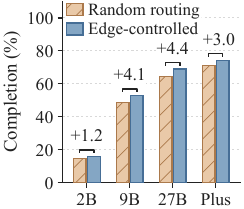}
        \caption{Cloud-invocation decision making.}
        \label{fig:rq3_a}
    \end{subfigure}
    \hfill
    \begin{subfigure}[t]{0.46\linewidth}
        \centering
        \includegraphics[width=\linewidth]{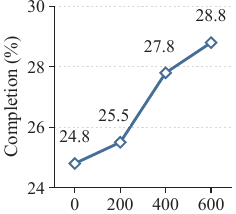}
        \caption{Guidance following.}
        \label{fig:rq3_b}
    \end{subfigure}
    
    \caption{
    Analysis of edge-side capabilities for increasingly edge-led collaboration.
    (a) Task completion under agent-controlled invocation and a matched random baseline with the same cloud-call rate.
    (b) \bench{} task completion of the edge model under cloud advisory during instruction-following training.
    }
    \label{fig:rq3}
    \vspace{-1.2em}
\end{wrapfigure}

%% file: sections/6_conclusion.tex
\section{Conclusion}

% In this work, we study edge--cloud collaboration for long-horizon LLM agent execution, where cloud participation is interleaved throughout the execution trajectory. To enable systematic investigation of this problem, we introduce \bench{}, an evaluation framework that combines executable digital-workspace agent tasks with unified measurements of task performance, resource usage, and privacy exposure. Our findings show that edge--cloud collaboration can achieve a favorable balance between capability, cost, and privacy compared with purely edge-side or cloud-side execution. As edge models become stronger, collaborative execution increasingly approaches cloud-only performance, while privacy exposure remains highly dependent on how cloud participation is organized. We further find that stronger edge models benefit from agent-controlled cloud invocation and advisory interaction. Further analyses suggest that effective cloud-invocation decisions and instruction-following are important abilities for effective edge--cloud collaboration, motivating future work to optimize edge--cloud collaboration的能力 for 形成 more capable edge agents.

In this work, we study edge--cloud collaboration for long-horizon LLM agent execution, where cloud participation is interleaved throughout the execution trajectory. To enable systematic investigation of this problem, we introduce \bench{}, an evaluation framework that combines executable digital-workspace agent tasks with unified measurements of task performance, resource usage, and privacy exposure. Our findings show that edge--cloud collaboration can achieve a favorable balance between capability, cost, and privacy compared with purely edge-side or cloud-side execution. As edge models become stronger, collaborative execution increasingly approaches cloud-only performance, while privacy exposure remains highly dependent on how cloud participation is organized. We further find that stronger edge models benefit from agent-controlled cloud invocation and advisory interaction. Additional analyses suggest that effective cloud-invocation decision-making and instruction-following are important abilities for edge--cloud collaboration, motivating future work to develop more capable edge agents by strengthening their ability to collaborate with cloud models.

\section*{AI Use Statement}
Generative AI tools were used to assist with language polishing and improving the clarity of the manuscript. All research ideas, experimental design, implementation, analysis, and conclusions were developed and verified by the authors, who take full responsibility for the content of the paper.
% No generative AI tools were used in the preparation of this manuscript.

\section*{Ethics Statement}
This work studies edge--cloud collaboration for LLM-powered agents and evaluates different collaboration strategies in terms of task performance, resource usage, and privacy exposure. \bench{} is designed for research evaluation and responsible system development. We do not use real private user data to construct privacy-sensitive benchmark content. When additional sensitive information is required for evaluation, it is synthetically generated and does not correspond to real individuals, organizations, credentials, or financial records. Human annotation is used only to validate benchmark annotations and synthetic content; no annotator personal information is collected or analyzed. The privacy annotations in \bench{} are intended to support systematic evaluation of sensitive information exposed during edge--cloud collaboration and to facilitate the development of more privacy-aware agent systems.

\section*{Reproducibility Statement}
We provide detailed descriptions of the benchmark construction, collaboration strategies, experimental setup, and evaluation procedures in the main paper and appendix. In particular, we document the task adaptation and filtering process, sensitivity annotation procedure and validation, implementations of the four edge--cloud collaboration strategies, model and inference configurations, and the metrics used to evaluate task performance, cloud resource usage, edge computation, and privacy exposure. We further report the experimental settings and implementation details for the analyses conducted in each research question. These details are provided to facilitate reproduction of the benchmark and the reported experimental results.

% 道德statement

% limitation

%% file: sections/n_appendix.tex
\clearpage
\appendix

\section{Appendix}
\label{sec:appendix}

\subsection{License}
\label{app:license}

All source benchmarks incorporated into \bench{} are used under their respective licenses. Claw-Eval, WildClawBench, QwenClawBench, LiveClawBench, and PinchBench are released under the MIT License, while ClawBench is released under the Apache-2.0 License. We preserve attribution to the original benchmarks for all adapted tasks and follow the redistribution and usage requirements specified by their licenses.

\subsection{Detailed Comparison with Prior Benchmarks}
\label{app:benchmark_comparison}

We provide additional details on how \bench{} differs from representative benchmarks for LLM routing and edge--cloud collaboration. Overall, prior benchmarks primarily study model allocation over queries or conversational turns and evaluate task quality together with inference-related costs, whereas \bench{} extends the evaluation to long-horizon agent execution, broader collaboration mechanisms, and trajectory-level privacy exposure. We next compare these benchmarks along three aspects: evaluation scenario, collaboration mode, and evaluation dimension.

\textbf{Evaluation Scenario.}
Existing benchmarks mainly evaluate routing over conventional LLM queries or conversational interactions, while \bench{} focuses on long-horizon agent execution in interactive environments. RouterBench~\citep{hu2024routerbench} evaluates routing across a collection of general LLM tasks. LLMRouterBench~\citep{li2026llmrouterbench} further broadens the task coverage to include software-engineering and tool-use benchmarks in addition to conventional reasoning, knowledge, and code-generation tasks. LLMSwitchBench~\citep{liu2025llmswitchbench} focuses on independent smart-home queries spanning binary, multiple-choice, and open-ended formats. M4A1~\citep{multi2026device} considers multi-modal and multi-task interactions within multi-turn conversations. In comparison, \bench{} evaluates collaboration over multi-step agent trajectories, where models repeatedly interact with tools, applications, and persistent workspace states.

\textbf{Collaboration Modes.}
Prior benchmarks predominantly study routing-based collaboration, whereas \bench{} additionally considers collaboration in which the edge agent itself controls cloud participation. RouterBench, LLMRouterBench, and LLMSwitchBench formulate model allocation primarily as query routing, where each input query is assigned to a selected model. M4A1 allows device--cloud placement to be reconsidered for different tasks throughout a multi-turn conversation. \bench{} includes both Query Routing and Per-Step Routing, and further considers two edge-controlled modes. Edge-Controlled Cloud Delegation allows the edge agent to decide when to delegate the next execution step to the cloud, while Edge-Controlled Cloud Advisory lets the edge agent request cloud guidance while retaining responsibility for subsequent action generation.

\textbf{Evaluation Dimensions.}
Existing benchmarks primarily evaluate task quality and inference-related efficiency, while \bench{} additionally quantifies privacy exposure over the complete execution trajectory. RouterBench focuses on task performance and economic cost, while LLMRouterBench evaluates performance-oriented and performance--cost routing and additionally provides latency-aware analysis. M4A1 jointly considers response quality, latency, and usage cost in device--cloud offloading. LLMSwitchBench incorporates privacy-sensitive queries and privacy-aware routing mechanisms, including local execution and masking before cloud inference, while its main routing evaluation reports task-quality metrics and inference cost. In \bench{}, privacy exposure is directly quantified by tracking which annotated sensitive information from the local workspace becomes visible to the cloud during execution. Resource usage is likewise measured over the complete trajectory, including cloud-side token consumption and monetary cost as well as edge-side computation.

\subsection{Task Suite Construction and Statistics}
\label{app:task_statistics}

This section provides additional details on the construction and composition of the \bench{} task suite, including collaboration-oriented difficulty filtering, coverage supplementation, and task statistics.

We first collect candidate tasks from multiple Claw-style executable-workspace agent benchmarks and adapt them to the \bench{} execution environment following the protocol in Appendix~\ref{app:task_adaptation}. We then perform collaboration-oriented difficulty filtering to retain tasks that provide meaningful room for edge--cloud collaboration. All filtering runs are conducted on the adapted task instances in the standardized \bench{} environment, using the same workspace states, tool interfaces, and task-specific evaluation checks defined after adaptation. Specifically, each task is evaluated under three endpoint configurations, using Qwen3.5-2B and Qwen3.5-9B as representative edge models and GLM-5.3 as the cloud model. We use the resulting task completion scores as the filtering signal to characterize whether a task is already handled by edge-only execution, can be completed by cloud-only execution, or remains challenging for both endpoints.

\input{tables/data_source}

We remove tasks that are consistently completed by both edge-only and cloud-only execution, since such tasks provide limited opportunity to differentiate execution paradigms. We also exclude tasks where neither endpoint achieves meaningful completion, since such tasks provide limited basis for evaluating collaborative execution. The resulting task suite focuses on scenarios where edge-only, cloud-only, and collaborative execution can exhibit meaningful differences. Importantly, this criterion does not require all edge models to fail: tasks that are challenging for Qwen3.5-2B but can be handled by Qwen3.5-9B are retained, as they remain informative for studying how collaboration evolves with increasing edge capability.

\input{tables/task_cat}

After endpoint filtering, we review the remaining task pool for coverage across task types, workspace workflows, and agentic operations. Following the operation taxonomy used in prior work~\citep{zai2025zclawbench}, we identify underrepresented categories and construct a small number of supplementary tasks to improve coverage. For each target category, human authors define the task scenario, objective, required agentic operations, and evaluation criteria. Claude Opus 4.8 is used to assist with the implementation of the corresponding workspace artifacts, application interfaces, and executable evaluation checks. Each constructed task is then manually reviewed for consistency among the task objective, workspace setup, required interactions, and evaluation checks, and executed to verify that the environment and checks function as intended. Instances that fail validation are discarded. This process results in the final suite of 128 executable tasks. Table~\ref{tab:task_source_stats} summarizes the task sources, while Table~\ref{tab:task_category_stats} reports the distribution across agentic operation categories, showing broad coverage of digital-workspace workflows.

% and task adaptation procedure
% 数据改造过程描述
\subsection{Task Adaptation Protocol}
\label{app:task_adaptation}

This section describes how tasks collected from heterogeneous executable-workspace benchmarks are adapted into standardized executable instances for \bench{}. The adaptation protocol consists of three components: standardizing task execution, standardizing task evaluation, and validating that the resulting instances preserve comparable task difficulty. We then illustrate the resulting task format and execution workflow with a representative \bench{} instance.

\textbf{Standardization and Fidelity Check.}
Inspired by the task format used in~\citet{ding2026wildclawbench}, we first standardize task execution by organizing each task into a unified configuration and workspace structure. Each task is converted into a unified configuration format specifying its prompt, workspace paths, tools, environment settings, and other execution requirements. Task-relevant files and resources are organized into a common workspace structure, while tool and skill settings are aligned across tasks so that instances from different sources can follow the same agent loop and logging mechanism. We next standardize task evaluation while preserving the original success criteria of each task. The original verification logic is converted into a set of task-specific evaluation checks compatible with \bench{}, including rule-based validation, file or application-state auditing, command-output inspection, and rubric-based LLM-as-a-judge verification when semantic judgment is required. The resulting check scores are aggregated into the task completion score defined in Section~\ref{subsec:benchmark_evaluation}.

To validate adaptation fidelity, we examine whether the adapted instances preserve comparable execution and evaluation outcomes. For each task, we execute both the original and adapted versions with Qwen3.5-9B and compare their average completion scores over three independent runs. Tasks whose absolute difference in average completion score exceeds 0.3 are discarded. This validation helps identify substantial behavioral or scoring shifts introduced during adaptation.

\input{tables/example}

\textbf{Representative Task Workflow.}
To illustrate the resulting standardized task format, Table~\ref{tab:task_example} presents a representative \bench{} instance for customer follow-up, including its task instruction, expected behavior, evaluation checks, executable workspace, and available skills. The example requires the agent to identify customers overdue for contact, inspect their previous email communications, and prepare tier-specific follow-up drafts using CRM and Gmail services.

Using this task as an example, the harness first parses the standardized task configuration to obtain the task instruction, workspace, required skills, environment settings, initialization commands, and evaluation checks. It then creates an isolated execution environment for the task, loads the task workspace and task-specific skills, and executes the warmup procedure to instantiate the required application state. In this example, the customer and inbox data are used to initialize local CRM and Gmail services, which expose the application interfaces subsequently used by the agent. The evaluation ground truth is kept separate from the execution environment and is not made available to the agent during this stage. The harness then configures the evaluated model, starts the OpenClaw runtime, and passes the task instruction to the agent. During execution, the agent iteratively interacts with the CRM and Gmail services through the provided skills, observes the resulting application states, and decides subsequent actions until the task terminates or reaches the time limit. The complete interaction history is recorded as the execution trajectory and retained for subsequent analysis. After execution terminates, the ground-truth evaluation data are injected into the evaluation environment, and the task-specific checks assess customer identification, email-style differentiation, and draft quality to compute the task completion score. For privacy evaluation, the annotated sensitivity units are matched against the cloud-visible contexts recorded during collaborative execution to determine privacy exposure. The harness additionally collects the execution transcript and model-usage statistics for resource evaluation.

% Detailed annotation procedures, category definitions, statistics, and weighting details are provided in Appendix~\ref{app:privacy}.

\input{tables/privacy_cat}

\subsection{Privacy Annotation}
\label{app:privacy_annotation_construction}

\input{tables/apped_privacy_quality}

This subsection describes how sensitivity-unit annotations are constructed in \bench{}. Since privacy-relevant information in digital workspaces can be distributed across heterogeneous files, application states, and other task artifacts, we adopt an LLM-based annotation process to identify and annotate sensitive information from two sources: information already present in the task environment and realistically synthesized sensitive content introduced into suitable task artifacts. Across the resulting task environments, more than 15,000 information units are represented in total, of which 3,385 are annotated as sensitive, spanning 100 tasks and 12 fine-grained privacy categories. Both sources are represented under a common annotation schema and privacy taxonomy, with annotation quality further validated through human evaluation. The resulting sensitivity-unit annotations are used to measure privacy exposure during collaborative execution. Privacy exposure quantifies the fraction of annotated sensitive information units that become visible to the cloud through cloud-visible contexts along the execution trajectory.

\textbf{Identification of Pre-existing Sensitive Information.}
For each task, we use an LLM to inspect the local workspace and identify privacy-relevant information. The workspace may contain files, application states, and other task artifacts accessible to the agent during execution. Each identified item is grounded in its source artifact or application record, and the corresponding sensitive content and source location are recorded as a sensitivity unit. This grounding establishes both that the information is available within the task environment and where it originates, providing a concrete reference for subsequent exposure evaluation. The resulting annotations are stored separately as task metadata.

\textbf{Synthesis of Sensitive Information.}
To broaden the coverage of sensitivity annotations across tasks and privacy categories, we additionally introduce synthetic sensitive content into suitable task artifacts. We only augment tasks for which such content can be incorporated without changing the task objective, required interaction, expected output, or success criteria. For each eligible task, an LLM-assisted pipeline selects an appropriate artifact and insertion location and generates synthetic sensitive content consistent with the surrounding format and semantics. Each inserted span is then recorded as a sensitivity unit using the same annotation schema as pre-existing sensitive information.

We further apply a separate LLM-based verification step to improve the realism of the synthesized content. The verifier examines whether each synthetic unit resembles plausible real-world sensitive information and replaces unrealistic or placeholder-like values with more realistic synthetic alternatives. Table~\ref{tab:privacy_verification_examples} presents representative examples before and after verification.

\textbf{Sensitivity-Unit Representation and Taxonomy.}
All sensitivity units, whether identified from the original workspace or synthetically introduced, follow a common annotation schema that records the sensitive content, its source location, and its privacy category. The source location links each unit to the artifact or application record from which it originates, allowing its exposure to be traced during agent execution. We organize sensitivity units into two top-level domains---\textit{Personal Sensitive Information} and \textit{Organizational Confidential Information}---comprising 12 fine-grained categories in total. The personal-information categories build on prior privacy taxonomies~\citep{ngong2026agentscope,shen2025pii}, while the organizational categories are further informed by WIPO guidance on trade secrets~\citep{WIPO2024TradeSecrets}. Table~\ref{tab:privacy_annotation_statistics} summarizes the resulting taxonomy, annotation statistics, and task coverage.

We use GLM-5.3 for sensitivity identification and synthesis, while GPT-5.5 serves as an independent verifier for the synthesized content. We further conduct human evaluation to validate the accuracy of pre-existing sensitivity-unit identification and the realism and placement quality of synthetic sensitive content. Detailed protocols and results are provided in Appendix~\ref{app:privacy_annotation_validation}.

\subsection{Validation of Privacy-Exposure Detection}
\label{app:privacy_leakage_validation}

We validate privacy-exposure detection in two stages. We first measure the consistency of human exposure judgments to establish reliable reference labels, and then evaluate the automated detector against the adjudicated human labels. We conduct this validation on 200 sensitivity-unit--context pairs sampled from benchmark trajectories.

\input{tables/append_llm_judge_verify}

Specifically, for each pair, two trained annotators are provided with the sensitivity unit and the corresponding cloud-visible context and asked to determine whether the information represented by the unit is disclosed to the cloud. The annotators are blinded to both the automated detector's prediction and the identity of the evaluated agent and independently assign a binary exposure label. Disagreements are resolved by a third annotator to obtain an adjudicated human reference label. We report raw inter-annotator agreement and Cohen's $\kappa$~\citep{cohen1960coefficient} to measure human consistency, and evaluate GPT-5.4-mini against the adjudicated human labels using precision, recall, and F1 score.

As shown in Table~\ref{tab:leakage_judge_validation}, the two human annotators achieve 97.5\% raw agreement and a Cohen's $\kappa$ of 0.945, indicating highly consistent exposure judgments. Against the adjudicated human labels, GPT-5.4-mini achieves 97.7\% precision, 96.2\% recall, and 96.9\% F1, demonstrating close agreement with human judgments.
% 人工评价

% \subsection{Human Validation of Privacy Evaluation}
% \label{append:valid_the_llm_judge}
% To assess the quality and reliability of the privacy evaluation in \bench{}, we conduct three human evaluation studies: validation of privacy leakage judgments, reliability assessment of sensitivity-unit annotations, and quality assessment of synthetically generated sensitive content. Across all three studies, we recruit three trained annotators with experience in NLP research. Before formal annotation, they receive task-specific guidelines and complete a pilot round to calibrate their understanding of the evaluation criteria. The pilot samples are excluded from the reported results, and all initial annotations are completed independently.

\subsection{Validation of Privacy Annotations}
\label{app:privacy_annotation_validation}

We validate the two components of privacy annotation in \bench{} through complementary human studies: the accuracy of sensitivity-unit identification in the original task environments and the quality of synthetically introduced sensitive content.

\input{tables/append_sensitivity_unit_validation}
\textbf{Sensitivity-Unit Identification Reliability.}
We first assess the reliability of the LLM-assisted identification of pre-existing sensitive information. We randomly sample 50 sensitivity units identified by GLM-5.3 and ask a human annotator to determine whether each unit indeed contains privacy-relevant information based on its source artifact and surrounding context. We report the proportion of identified units validated as sensitive by the human annotator as annotation precision.

As shown in Table~\ref{tab:sensitivity_unit_validation}, 45 of the 50 sampled units are validated as privacy-relevant, yielding an annotation precision of 90.0\%. This result indicates that the identified sensitivity units are generally reliable.

\input{tables/append_privacy_validate}
\textbf{Synthetic Content Quality.}
We further assess whether the synthetically introduced sensitive content is realistic and naturally integrated into its surrounding task context. We sample 150 synthetic sensitivity units from 50 tasks and ask two human annotators to independently evaluate each unit along two dimensions: \textit{content realism} and \textit{placement naturalness}. Using a five-point Likert scale~\citep{likert1932technique}, the annotators rate whether the sensitive information itself is plausible and whether it fits naturally within its specific workspace location and surrounding context. For each unit, the annotators are shown the task instruction and the relevant workspace context.

As shown in Table~\ref{tab:human_validation}, the synthetic sensitivity units receive average ratings of 4.62 for content realism and 4.59 for placement naturalness. Quadratically weighted Gwet's AC2~\citep{gwet2008computing} reaches 0.91 and 0.90 for the two dimensions, respectively, indicating high inter-annotator agreement. These results show that the synthesized sensitive content is both realistic and naturally integrated into the task environment.

% \subsection{Generalization to a Different Edge Model Family}
% \label{app:gemma_generalization}

% To examine whether the observed collaboration behavior generalizes beyond the Qwen model family used in our main experiments, we conduct an additional evaluation using Gemma-4-E2B-it as the edge model. We retain GLM-5.3 as the cloud model and evaluate the same four collaboration strategies---Query Routing, Per-Step Routing, Edge-Controlled Cloud Delegation, and Edge-Controlled Cloud Advisory---under the same benchmark and evaluation protocol. All other execution and evaluation settings remain unchanged.

% \input{tables/append_gemma4}

% As shown in Table~\ref{tab:gemma4_results}, edge--cloud collaboration consistently improves task performance over edge-only execution with Gemma-4-E2B-it, while avoiding the resource usage and privacy exposure of cloud-only execution. Moreover, the relative behavior of different collaboration strategies remains broadly consistent with the observations from the Qwen-based experiments: different strategies can achieve similar task performance while exhibiting substantially different cloud usage and privacy exposure. These results suggest that the main findings of \bench{} are not specific to the Qwen model family and extend to a different edge-model architecture.

\subsection{Additional Evaluation with Alternative Cloud Models}

\label{app:deepseek_v41}
To further examine the generality of our findings beyond the primary cloud model used in the main experiments, we conduct additional experiments using DeepSeek-V4.1-Flash~\citep{deepseekai2026deepseekv4} as an alternative cloud model. DeepSeek-V4.1-Flash is selected due to its similar capability level to the primary cloud model, allowing us to evaluate whether our observations hold under a different cloud model. All experimental settings, including the task suite, edge models, collaboration strategies, evaluation metrics, and execution environment, remain identical to those in the main experiments.

Table~\ref{tab:deepseek_v41} reports the results. The additional experiments show consistent trends with the main results: edge--cloud collaboration provides a favorable balance between task performance, resource usage, and privacy exposure compared with edge-only and cloud-only execution, while the effectiveness of different collaboration strategies varies with edge-model capability. These results further demonstrate that our observations are not specific to a single cloud model.

\input{tables/dpskv41}

\subsection{Implementation Details}
\label{app:baselines}
This section provides implementation details for reproducing the experiments in \bench{}, including the execution environment and the implementation of the evaluated edge--cloud collaboration strategies.

\textit{Experimental Environment.}
All experiments are conducted using OpenClaw as the agent execution framework. The maximum context window is configured to 128K tokens, with each individual model interaction limited to 32K tokens. OpenClaw additionally employs an automatic context compression mechanism to manage context growth during long-horizon execution. The agent execution environment is hosted in E2B sandboxes with stable resource allocation to ensure consistent and reliable execution across experiments. All collaboration strategies share the same agent runtime, execution environment, and model deployment configuration. For all experiments, when cloud assistance is requested, $v_t^c$ consists of the current task instruction and the execution history maintained by OpenClaw, including previous actions, tool calls, and environment observations. No additional redaction or privacy filtering is applied before cloud invocation.

\textit{Query Routing.}
Query Routing makes a single routing decision before agent execution and assigns the entire task trajectory to either the edge model $M_e$ or the cloud model $M_c$. We directly use the matrix-factorization-based router released with RouteLLM~\citep{ong2024routellm}. We calibrate the threshold on the validation set to obtain an approximately 15\% cloud-participation rate at the step level and apply the same threshold during evaluation. This target cloud-participation rate is selected to keep the cloud usage of routing-based strategies comparable to that of edge-controlled strategies, enabling a more controlled comparison under similar cloud resource budgets.

\textit{Per-Step Routing.}
Per-Step Routing revisits the edge--cloud decision at every agent interaction step. We adapt the uncertainty signal used by GlimpRouter~\citep{zeng2026glimprouter} to agent execution. Before generating the action at each step, we obtain the edge model's next reasoning-token distribution and compute its entropy as an uncertainty score. If the entropy exceeds a calibrated threshold, the current step is routed to $M_c$ for action generation; otherwise, generation continues with $M_e$. The routing decision is repeated independently throughout the trajectory. We calibrate the routing threshold on the validation set such that approximately 15\% of tasks are assigned to the cloud, aiming to maintain a cloud usage level comparable to other collaboration strategies, and apply the same threshold during evaluation.

\textit{Edge-Controlled Cloud Delegation.}
Unlike the two routing-based strategies above, Edge-Controlled Cloud Delegation places the cloud-participation decision within the edge agent itself. The edge model is provided with a delegation tool backed by $M_c$ and can invoke it at any point during execution. When invoked, the current conversation history, including previous actions, tool calls, and observations, is forwarded to the cloud model, which directly performs the next agent step. Control then returns to the edge model for subsequent execution. The delegation instruction encourages the edge agent to invoke the cloud before committing to a substantive approach, when it encounters difficulty or considers changing its approach, and before declaring the task complete. Thus, both the timing and frequency of cloud participation are determined dynamically by the edge agent rather than by an external router.

\textit{Edge-Controlled Cloud Advisory.}
Edge-Controlled Cloud Advisory uses the same agent-controlled invocation mechanism but assigns a different role to the cloud model. The edge model is provided with an advisory tool and autonomously decides when cloud guidance is needed during execution. Upon invocation, the current execution history, including previous actions, tool calls, and environment observations, is provided to $M_c$. The cloud model returns advisory information in natural language, such as plans, suggestions, or corrective feedback. The advice is appended to the edge model's context, after which $M_e$ generates the next action based on the task instruction, execution history, and received advice.

\input{figs/figs/if}

\subsection{Understanding Edge-Side Abilities for Collaboration}
\label{app:rq3_analysis}

To further understand the edge-side abilities associated with the observed shift toward edge-controlled collaboration, we conduct two controlled analyses focusing on cloud-assistance decision making and guidance following.

\textbf{Cloud-Assistance Decision Making.}
Directly evaluating the quality of cloud-invocation decisions is challenging because there is no explicit ground-truth annotation indicating when cloud assistance is necessary during agent execution. Therefore, we evaluate the effectiveness of edge-controlled invocation through a matched random baseline. For each edge model, we first execute the Edge-Controlled Cloud Delegation strategy and record its cloud-call rate. We then construct a random-invocation baseline that preserves the same cloud-call rate but randomly selects the invocation steps throughout the trajectory. Both approaches therefore receive the same amount of cloud assistance, while differing only in whether the invocation timing is determined by the edge agent or randomly selected. We compare their task completion performance, where the performance gain over random invocation serves as an indicator of whether the edge agent can identify execution states where cloud assistance is beneficial.

\textbf{Cloud Guidance Utilization.}
To examine whether instruction-following ability is associated with effective advisory collaboration, we conduct a controlled instruction-following training experiment. Starting from Qwen3.5-2B, we apply targeted instruction-following RL training using the IF data provided by Ultradata~\citep{wang2026datasciencetechnologyagi}. The model is trained for 600 steps with GRPO, using a batch size of 32 per step and 8 rollouts per prompt.

To separate guidance utilization from cloud-invocation decisions, we evaluate all checkpoints under a fixed advisory setting, where cloud advisory calls are triggered at predetermined edge execution steps (e.g., every five steps) rather than dynamically selected by the edge model. We evaluate the model from two perspectives. First, we measure instruction-following ability using three instruction-following benchmarks (IFEval~\citep{zhou2023instruction}, IFBench~\citep{pyatkin2026generalizing}, and Multi-IF~\citep{he2024multi}) to verify that the training improves the targeted capability. Second, we evaluate the resulting models on \bench{} under both advisory collaboration and edge-only execution. The edge-only setting serves as a control to examine whether improvements are accompanied by general gains in agent execution capability.

As shown in Figure~\ref{fig:if}, instruction-following performance consistently improves throughout training, while edge-only task performance remains largely unchanged. In contrast, advisory collaboration performance increases with instruction-following training, improving from 24.8\% at initialization to 28.8\% after 600 training steps. These results suggest that instruction-following ability is particularly relevant to cloud guidance utilization, beyond general improvements in edge-only task execution.

\subsection{Pricing and FLOPs Estimation}
\label{app:pricing}

This section describes how cloud-side monetary cost and edge-side computation are estimated in \bench{}.

For each cloud invocation, the harness records raw input tokens, cache-read input tokens, and output tokens. We aggregate these token counts over all cloud invocations and compute the corresponding monetary cost as:
\begin{equation}
\small
C_{\mathrm{cloud}}
=
\frac{
p_{\mathrm{raw}} T_{\mathrm{raw}}
+
p_{\mathrm{cache}} T_{\mathrm{cache}}
+
p_{\mathrm{out}} T_{\mathrm{out}}
}{10^{6}},
\end{equation}
where $T_{\mathrm{raw}}$, $T_{\mathrm{cache}}$, and $T_{\mathrm{out}}$ denote the aggregated raw input, cache-read input, and output token counts, respectively, and $p_{\mathrm{raw}}$, $p_{\mathrm{cache}}$, and $p_{\mathrm{out}}$ denote their corresponding prices per million tokens. Table~\ref{tab:cloud_pricing} reports the OpenRouter pricing used for cost estimation, as of August 2026.

\input{tables/price}

For edge-side execution, we use estimated FLOPs as a proxy for local computation. Following the parameter-based approximation of~\citet{kaplan2020scaling}, we approximate inference FLOPs as:
\begin{equation}
\small
\mathrm{FLOPs}_{\mathrm{edge}}
=
2 \cdot N_{\mathrm{act}} \cdot T_{\mathrm{edge}},
\end{equation}
where $N_{\mathrm{act}}$ denotes the number of parameters activated per token and $T_{\mathrm{edge}}$ is the total number of input and output tokens processed by the edge model across all invocations. For dense models, $N_{\mathrm{act}}$ corresponds to the total parameter count, while for mixture-of-experts models, it corresponds to the parameters activated for each token.

%% file: tables/data_source.tex
% \begin{table}[h]
\begin{wraptable}{r}{0.46\textwidth}
\vspace{-1em}
\centering
\small
\caption{Source distribution of the \bench{}.}
\label{tab:task_source_stats}
\vspace{-0.5em}
\begin{tabular}{lcc}
\toprule
\textbf{Task Source} & \textbf{\# Tasks} & \textbf{Percentage} \\
\midrule
Claw-Eval & 68 & 53.12\% \\
WildClawBench & 22 & 17.19\% \\
QwenClawBench & 21 & 16.41\% \\
LiveClawBench & 4 & 3.12\% \\
PinchBench & 3 & 2.34\% \\
ClawBench & 3 & 2.34\% \\
Curated Tasks & 7 & 5.47\% \\
\midrule
Total & 128 & 100.00\% \\
\bottomrule
\end{tabular}

% \vspace{-1em}
\end{wraptable}

% \end{table}

%% file: tables/task_cat.tex
% \begin{table}[h]

\begin{wraptable}{r}{0.46\textwidth}
\vspace{-1em}
\centering
\small
\caption{Distribution of \bench{} tasks across agentic operation categories.}
\label{tab:task_category_stats}
\vspace{-0.5em}
\setlength{\tabcolsep}{5pt}
\begin{tabular}{lr}
\toprule
\textbf{Task Category} & \textbf{\# Tasks} \\
\midrule
Office \& Daily Tasks & 36 \\
Information Search \& Gathering & 34 \\
Safety \& Security  & 21 \\
Development \& Operations & 14 \\
Data Analysis & 13 \\
Automation & 10 \\
\midrule
Total & 128 \\
\bottomrule
\end{tabular}

% \end{table}
\end{wraptable}

%% file: tables/example.tex
\begin{table*}[t]
\centering
\small
\caption{
Representative standardized task instance in \bench{}.
The example illustrates the task specification, evaluation criteria, and executable environment used during agent execution.
For readability, some fields are shortened and reformatted for presentation while preserving their original semantics.
}
\label{tab:task_example}
\begin{tabular}{@{}p{0.17\textwidth}p{0.79\textwidth}@{}}
\toprule
\textbf{Component} & \textbf{Representative Example} \\
\midrule

\multicolumn{2}{l}{\textbf{\textit{Task Specification}}} \\
\midrule

\textbf{Task Instruction}
&
Identify active customers who have not been contacted for more than 30 days, review their previous email communications, and draft follow-up emails with different styles for VIP and standard customers while excluding churned customers.
\\

\textbf{Expected Behavior}
&
Query the CRM to identify overdue active customers, exclude churned customers, retrieve relevant email records to recover previous communication context, and prepare personalized follow-up drafts according to customer tier and communication history.
\\

\midrule
\multicolumn{2}{l}{\textbf{\textit{Evaluation}}} \\
\midrule

\textbf{Evaluation Checks}
&
Customer identification (0.35), differentiation between VIP and standard follow-up styles (0.35), and follow-up draft quality (0.30). The weighted check scores are aggregated into the task completion score.
\\

\textbf{Sensitivity Annotation}
&
\texttt{sensitive\_items.json} contains fine-grained sensitivity units associated with sensitive information in the task environment and is used for privacy-exposure evaluation.
\\

\midrule
\multicolumn{2}{l}{\textbf{\textit{Execution Environment}}} \\
\midrule

\textbf{Workspace}
&
\begin{minipage}[t]{0.76\textwidth}
\vspace{0.2em}
\ttfamily
task\_65\_customer\_followup/\\
\quad exec/\\
\qquad mock\_services/\\
\qquad\quad crm/\\
\qquad\qquad server.py\\
\qquad\quad gmail/\\
\qquad\qquad server.py\\
\quad tmp/\\
\qquad customers.json\\
\qquad inbox.json\\
\quad gt/\\
\qquad sensitive\_items.json
\vspace{0.2em}
\end{minipage}
\\

\textbf{Skills}
&
CRM operations support listing customers and retrieving customer details, while Gmail operations support listing and retrieving messages, saving email drafts, and sending emails.
\\

\textbf{Environment Initialization}
&
Before agent execution, the customer and inbox data are used to initialize the task-specific CRM and Gmail services during warmup, providing the application state and interfaces required for subsequent interaction.
\\

\bottomrule
\end{tabular}
\end{table*}

%% file: tables/privacy_cat.tex
\begin{table*}[t]
\centering
\small
\caption{
Statistics of sensitivity-unit annotations by privacy domain and fine-grained category.
\# Tasks denotes the number of tasks containing at least one unit from that category.
}
\label{tab:privacy_annotation_statistics}
\vspace{-0.5em}

\begin{tabular}{llrrr}
\toprule
\textbf{Domain} & \textbf{Category} & \textbf{\# Units} & \textbf{Percentage} & \textbf{\# Tasks} \\
\midrule
\multirow{5}{*}{\makecell[l]{Personal Sensitive\\Information}}
& P1 Identity identifier & 276 & 8.15\% & 48 \\
& P2 Documents and accounts & 457 & 13.50\% & 66 \\
& P3 Contact information & 1178 & 34.80\% & 84 \\
& P4 Financial information & 209 & 6.17\% & 48 \\
& P5 Sensitive attributes & 248 & 7.33\% & 26 \\
\cmidrule{2-5}
& \textit{Subtotal} & \textbf{2368} & \textbf{69.96\%} & -- \\
\midrule
\multirow{7}{*}{\makecell[l]{Organizational\\Confidential\\Information}}
& S1 Code or algorithm & 6 & 0.18\% & 5 \\
& S2 Credentials or keys & 196 & 5.79\% & 32 \\
& S3 Security vulnerability & 26 & 0.77\% & 11 \\
& S4 Business strategy & 235 & 6.94\% & 54 \\
& S5 Personnel or administration & 97 & 2.87\% & 33 \\
& S6 Customer or partner information & 189 & 5.58\% & 46 \\
& S7 Internal finance & 268 & 7.92\% & 52 \\
\cmidrule{2-5}
& \textit{Subtotal} & \textbf{1017} & \textbf{30.04\%} & -- \\
\midrule
\textbf{Total} & -- & \textbf{3385} & \textbf{100.00\%} & -- \\
\bottomrule
\end{tabular}

\vspace{-0.5em}
\end{table*}

%% file: tables/apped_privacy_quality.tex
\begin{table*}[t]
\centering
\small
\caption{
Representative revisions of synthetic sensitivity units by the verifier.
Values are shortened where necessary for readability, and all examples are synthetic.
}
\label{tab:privacy_verification_examples}
\vspace{-0.5em}
\begin{tabular}{@{}llll@{}}
\toprule
\textbf{Type} & \textbf{Before} & \textbf{After} & \textbf{Identified Issue} \\
\midrule
Phone number
& \texttt{(415) 555-0298}
& \texttt{(415) 672-3845}
& Common fictional-number pattern. \\

Passport ID
& \texttt{E12345678}
& \texttt{E47291856}
& Obvious sequential placeholder pattern. \\

Cloud access key
& \texttt{AKIA...EXAMPLE}
& \texttt{AKIA...Z1R6}
& Contains a recognizable example token. \\

Password
& \texttt{password}
& \texttt{xK9\#...Q4wR}
& Common default or placeholder value. \\
\bottomrule
\end{tabular}
\end{table*}

%% file: tables/append_llm_judge_verify.tex
\begin{wraptable}{r}{0.6\textwidth}
    \centering
    \small
    \vspace{-1.3em}
    \caption{Validation results for privacy-exposure detection.}
    \label{tab:leakage_judge_validation}
    \vspace{-0.5em}
    \begin{tabular}{lccc}
        \toprule
        & \textbf{Agreement} & \textbf{Cohen's $\kappa$} & \textbf{$N$} \\
        \midrule
        Human annotators & 97.5\% & 0.945 & 200 \\
        \midrule
        & \textbf{Precision} & \textbf{Recall} & \textbf{F1} \\
        \midrule
        GPT-5.4-mini & 97.7\% & 96.2\% & 96.9\% \\
        \bottomrule
    \end{tabular}
    \vspace{-0.5em}
\end{wraptable}

%% file: tables/append_sensitivity_unit_validation.tex
\begin{wraptable}{r}{0.5\textwidth}
    \centering
    \small
    \vspace{-1.3em}
    \caption{Human validation of identified sensitivity units.}
    \label{tab:sensitivity_unit_validation}
    \vspace{-0.5em}
    \begin{tabular}{lccc}
        \toprule
        \textbf{Model}
        & \textbf{Valid}
        & \textbf{Invalid}
        & \textbf{Precision} \\
        \midrule
        GLM-5.3
        & 45
        & 5
        & 90.0\% \\
        \bottomrule
    \end{tabular}
    \vspace{-1em}
\end{wraptable}

%% file: tables/append_privacy_validate.tex
% \begin{table}[t]
\begin{wraptable}{r}{0.5\textwidth}
    \centering
    \small
    \vspace{-1.3em}
        \caption{
Human evaluation of 150 synthetic sensitivity units sampled from 50 tasks. Two annotators independently assess content realism and placement naturalness on a five-point Likert scale.
}
    \label{tab:human_validation}
    \vspace{-0.5em}
    \setlength{\tabcolsep}{5pt}
    \begin{tabular}{lcc}
        \toprule
        \textbf{Metric}
        & \textbf{Content}
        & \textbf{Placement} \\
        \midrule
        Mean rating
& 4.62
& 4.59 \\
Ratings $\geq 4$
& 86.3\%
& 88.3\% \\
Gwet's AC2
& 0.91
& 0.90 \\
        \bottomrule
    \end{tabular}
\vspace{-1em}
\end{wraptable}
% \end{table}

%% file: tables/dpskv41.tex
\definecolor{headblue}{HTML}{B7DDD9}
\definecolor{preexec}{HTML}{F4F0FF}
\definecolor{runtime}{HTML}{FFF3E6}

\begin{table*}[!t]
\centering
\small
\caption{
Performance of edge--cloud collaboration strategies on \bench{}.
\textbf{Performance}: task completion and Pass$^3$.
\textbf{Resource Usage}: Cloud Tok. (R/C/O) reports raw input, cache-read input, and output tokens in millions; Cloud Cost is reported in USD; Edge FLOPs are reported in PetaFLOPs.
\textbf{Privacy Exposure}: percentage of sensitive information exposed to the cloud (lower is better).
\runtimebox{} and \preexecbox{} denote current-generation and higher-capability proxy edge models, respectively. For Qwen3.7-Plus, Edge FLOPs are not reported because its parameter count is unavailable.
}
\label{tab:deepseek_v41}
\vspace{-0.3em}
\resizebox{\textwidth}{!}{
\begin{tabular}{lcccccc}
\toprule
\multirow{2}{*}{\textbf{Method}} & \multicolumn{2}{c}{\textbf{Performance$\uparrow$}} & \multicolumn{3}{c}{\textbf{Resource Usage$\downarrow$}} & \multirow{2}{*}{\textbf{Privacy Exp.$\downarrow$}} \\
\cmidrule(lr){2-3} \cmidrule(lr){4-6}
& Completion & Pass$^3$ & Cloud Tok. (R/C/O) & Cloud Cost & Edge FLOPs & \\
\midrule

\rowcolor{headblue}\multicolumn{7}{c}{\textbf{\textit{Edge-only}}} \\
Qwen3.5-2B~(\citeyear{qwen35}) & 12.75 & 2.34 & -- & -- & 142.98 & 0.00 \\
Qwen3.5-9B~(\citeyear{qwen35}) & 39.12 & 9.38 & -- & -- & 121.64 & 0.00 \\
Qwen3.5-27B~(\citeyear{qwen35}) & 61.79 & 30.47 & -- & -- & 564.49 & 0.00 \\
Qwen3.7-Plus~(\citeyear{qwen35}) & 71.34 & 53.12 & -- & -- & -- & 0.00 \\
% Qwen3.8-Max~(\citeyear{qwen35}) & 70.76 & 45.31 & -- & -- & 1542.25 & 0.00 \\
\midrule

\rowcolor{headblue}\multicolumn{7}{c}{\textbf{\textit{Cloud-only}}} \\
DeepSeek-v4-Pro~(\citeyear{deepseekai2026deepseekv4}) & 72.90 & 53.12 & 5.81 / 50.20 / 1.60 & 16.41 & -- & 86.66 \\
% GLM-5.3~(\citeyear{openai2026gpt54}) & 82.70 & 71.88 & 13.52 / 65.76 / 4.64 & 56.42 & -- & 87.01 \\
Deepseek-v4.1-Flash~(\citeyear{deepseekai2026deepseekv4}) & 81.42 & 71.88 & 14.35 / 90.32 / 2.77 & 4.08 & -- & 90.85 \\
\midrule

\rowcolor{headblue}\multicolumn{7}{c}{\textbf{\textit{Edge-Cloud Collaboration}}} \\

\textit{\textbf{Qwen3.5-2B}} \\
\rowcolor{runtime}\quad Query Routing~(\citeyear{ong2024routellm}) 
& 23.46 & 13.28 & 2.60 / 8.96 / 0.38 & 0.64 & 139.06 & 13.08 \\

\rowcolor{runtime}\quad Per-Step Routing~(\citeyear{zeng2026glimprouter}) 
& 42.70 & 10.94 & 8.11 / 17.72 / 0.51 & 1.57 & 70.04 & 46.31 \\

\rowcolor{runtime}\quad Edge-Ctrl. Delegation~(\citeyear{farzaneh2026think}) 
& 19.20 & 2.34 & 1.00 / 0.74 / 0.10 & 0.21 & 145.46 & 9.52 \\

\rowcolor{runtime}\quad Edge-Ctrl. Advisory~(\citeyear{anthropic2026advisor}) 
& 15.53 & 3.12 & 0.70 / 0.88 / 0.19 & 0.22 & 178.10 & 7.87 \\

\midrule

\textit{\textbf{Qwen3.5-9B}} \\
\rowcolor{runtime}\quad Query Routing~(\citeyear{ong2024routellm}) 
& 45.66 & 21.88 & 2.08 / 11.69 / 0.35 & 0.56 & 196.51 & 13.03 \\

\rowcolor{runtime}\quad Per-Step Routing~(\citeyear{zeng2026glimprouter}) 
& 50.36 & 19.53 & 2.57 / 4.15 / 0.14 & 0.48 & 31.20 & 28.74 \\

\rowcolor{runtime}\quad Edge-Ctrl. Delegation~(\citeyear{farzaneh2026think}) 
& 51.79 & 17.19 & 1.66 / 1.17 / 0.29 & 0.43 & 204.07 & 21.18 \\

\rowcolor{runtime}\quad Edge-Ctrl. Advisory~(\citeyear{anthropic2026advisor}) 
& 56.00 & 24.22 & 2.82 / 2.50 / 0.62 & 0.80 & 340.08 & 39.25 \\
\midrule

\textit{\textbf{Qwen3.5-27B}} \\
\rowcolor{preexec}\quad Query Routing~(\citeyear{ong2024routellm}) 
& 63.51 & 39.84 & 3.68 / 28.34 / 0.52 & 0.95 & 601.08 & 13.07 \\

\rowcolor{preexec}\quad Per-Step Routing~(\citeyear{zeng2026glimprouter}) 
& 67.24 & 42.97 & 3.96 / 8.47 / 0.27 & 0.78 & 85.38 & 38.47 \\

\rowcolor{preexec}\quad Edge-Ctrl. Delegation~(\citeyear{farzaneh2026think}) 
& 67.11 & 41.41 & 2.34 / 1.45 / 0.47 & 0.63 & 605.25 & 46.19 \\

\rowcolor{preexec}\quad Edge-Ctrl. Advisory~(\citeyear{anthropic2026advisor}) 
& 69.50 & 46.09 & 1.73 / 1.31 / 0.55 & 0.59 & 636.49 & 49.02 \\

\midrule

\textit{\textbf{Qwen3.7-Plus}} \\
\rowcolor{preexec}\quad Query Routing~(\citeyear{ong2024routellm}) 
& 73.88 & 60.16 & 4.07 / 14.97 / 0.48 & 0.94 & -- & 12.67 \\

\rowcolor{preexec}\quad Per-Step Routing~(\citeyear{zeng2026glimprouter}) 
& 73.86 & 53.91 & 2.61 / 2.67 / 0.15 & 0.49 & -- & 55.37 \\

\rowcolor{preexec}\quad Edge-Ctrl. Delegation~(\citeyear{farzaneh2026think}) 
& 75.04 & 59.38 & 2.41 / 2.75 / 0.41 & 0.61 & -- & 44.54 \\

\rowcolor{preexec}\quad Edge-Ctrl. Advisory~(\citeyear{anthropic2026advisor}) 
& 75.31 & 61.72 & 1.99 / 1.11 / 0.52 & 0.61 & -- & 65.56 \\
% \textit{\textbf{Qwen3.8-Max}} \\
% \rowcolor{preexec}\quad Query Routing~(\citeyear{ong2024routellm}) & 80.03 & 70.31 & 2.31 / 11.23 / 0.75 & 9.43 & 2180.89 & 12.46 \\
% \rowcolor{preexec}\quad Per-Step Routing~(\citeyear{zeng2026glimprouter}) & -- & -- & -- & -- & -- & -- \\
% \rowcolor{preexec}\quad Edge-Ctrl. Delegation~(\citeyear{farzaneh2026think}) & 80.63 & 71.88 & 2.74 / 3.52 / 1.02 & 9.22 & 2171.17 & 80.89 \\
% \rowcolor{preexec}\quad Edge-Ctrl. Advisory~(\citeyear{anthropic2026advisor}) & 82.64 & 72.66 & 2.73 / 1.00 / 0.80 & 7.61 & 2315.85 & 77.54 \\

\bottomrule
\end{tabular}
}

\vspace{-1em}
\end{table*}

%% file: figs/figs/if.tex
\begin{figure}[t]
    \centering
    \vspace{-1em}
    \begin{subfigure}[t]{0.23\linewidth}
        \centering
        \includegraphics[width=\linewidth]{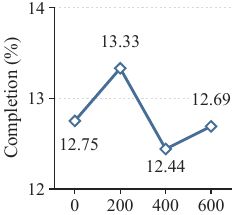}
        \caption{Edge-only performance on \bench{}.}
        \label{fig:if1}
    \end{subfigure}
    \hfill
    \begin{subfigure}[t]{0.23\linewidth}
        \centering
        \includegraphics[width=\linewidth]{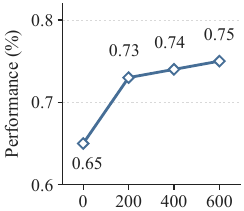}
        \caption{IFEval performance.}
        \label{fig:if2}
    \end{subfigure}
    \hfill
    \begin{subfigure}[t]{0.23\linewidth}
        \centering
        \includegraphics[width=\linewidth]{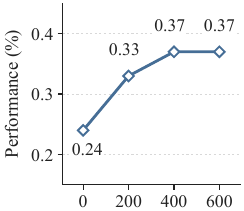}
        \caption{IFBench performance.}
        \label{fig:if3}
    \end{subfigure}
    \hfill
    \begin{subfigure}[t]{0.23\linewidth}
        \centering
        \includegraphics[width=\linewidth]{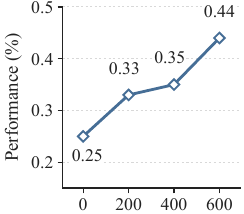}
        \caption{Multi-IF performance.}
        \label{fig:if4}
    \end{subfigure}
    
    \caption{
Training dynamics of instruction-following ability and edge-side task performance.
(a) Edge-only task performance on \bench{}.
(b)--(d) Instruction-following performance on three instruction-following benchmarks.
}
    \label{fig:if}
    \vspace{-1.5em}
\end{figure}

%% file: tables/price.tex
\begin{table}[t]
% \begin{wraptable}{r}{0.58\textwidth}
\centering
\small
% \vspace{-1.3em}
\caption{Cloud-side pricing used for monetary cost estimation. Prices are reported in USD per million tokens.}
\label{tab:cloud_pricing}
\vspace{-0.5em}
\begin{tabular}{lccc}
\toprule
\textbf{Cloud Model} & \textbf{Input} & \textbf{Cache Input} & \textbf{Output} \\
\midrule
DeepSeek-v4-Pro & 1.74 & 0.0145 & 3.48 \\
DeepSeek-v4.1-Flash & 0.15 & 0.003 & 0.6 \\
GLM-5.3 & 1.4 & 0.26 & 4.40 \\
\bottomrule
\end{tabular}
% \vspace{-1em}
% \end{wraptable}
\end{table}